\documentclass{article} % For LaTeX2e
\usepackage{iclr2027_conference,times}

\usepackage{amsmath,amsfonts,bm}

\def\eqref#1{equation~\ref{#1}}
\def\1{\bm{1}}

\DeclareMathAlphabet{\mathsfit}{\encodingdefault}{\sfdefault}{m}{sl}
\SetMathAlphabet{\mathsfit}{bold}{\encodingdefault}{\sfdefault}{bx}{n}

\usepackage{url}

\usepackage{times}
\usepackage{latexsym}
\usepackage{subcaption}
\usepackage[T1]{fontenc}
\usepackage[utf8]{inputenc}

\usepackage{microtype}

\usepackage{inconsolata}

\usepackage{graphicx}
\usepackage[colorlinks,linkcolor=black,citecolor=blue]{hyperref}
\usepackage{url}
\usepackage{booktabs}
\usepackage{amsfonts}
\usepackage{amsmath}
\usepackage{amssymb}
\usepackage{amsthm}
\usepackage{nicefrac}
\usepackage{microtype}
\usepackage[dvipsnames]{xcolor}
\usepackage{graphicx}
\usepackage{multirow}
\usepackage{colortbl}
\usepackage{enumitem}
\usepackage{algorithm}
\usepackage{algpseudocode}
\usepackage{pifont}
\usepackage{amsmath}
\usepackage{pifont}
\usepackage{tabularx}

\usepackage{subcaption}

\usepackage[table]{xcolor}

\definecolor{SpeedGreen}{RGB}{0,150,0}

\newcommand{\costone}{\text{\ding{182}}}
\newcommand{\costtwo}{\text{\ding{183}}}
\newcommand{\gdec}[1]{%
  \kern0.03em%
  \smash{\raisebox{-0.45ex}{%
    \textcolor{SpeedGreen}{\fontsize{7pt}{7pt}\selectfont$\downarrow$#1}%
  }}%
}

\definecolor{colOurs}{HTML}{EFF6FF}

\title{SpecFold: Folding Multi-Branch Redundancy for Faster Speculative Decoding in Diffusion Language Models}

\author{Chung-En~Ho, Weiyu~Sun, Cheng-Jhih~Shih, He Li, Yong Liu, Yingyan (Celine) Lin \\
Georgia Institute of Technology \\
\texttt{\{cho322,weiyu.lavender,cshih63,harvey-l,liuyong,ylin715\}@gatech.edu}
}

\iclrfinalcopy % Uncomment for camera-ready version, but NOT for submission.
\begin{document}

\maketitle

\begin{abstract}

Diffusion large language models (DLLMs) generate text through iterative block denoising, and multi-branch speculative decoding accelerates this process by verifying a main branch together with multiple draft branches in a single forward pass. While prior DLLM acceleration methods primarily exploit temporal redundancy across denoising steps, we identify a complementary redundancy axis within each speculative verification step: multi-branch computational redundancy. During speculative verification, draft branches inherit most tokens from their parents while unmasking a small set of additional positions, causing large portions of hidden states to remain highly similar across branches. We propose \textsc{SpecFold}, an algorithm--system co-design that exploits this multi-branch redundancy to reduce the cost of multi-branch speculative verification. Algorithmically, \textsc{SpecFold} performs token-level residual gating and selectively reuses parent computation through folded attention and FFN while preserving residual hidden states. Systemically, a Triton kernel implementation translates this fine-grained reuse into end-to-end throughput gains through efficient sparse multi-branch execution. \textsc{SpecFold} is orthogonal to temporal caching and compatible with existing DLLM speculation strategies. Across two DLLM families, five models, and five standard benchmarks, \textsc{SpecFold} achieves up to $1.64\times$ throughput over Spiffy and up to $1.99\times$ over vanilla decoding, while maintaining comparable task performance. Code is available at \url{https://github.com/chungen04/specfold}.

\end{abstract}

\section{Introduction}
\label{sec:intro}

Diffusion large language models (DLLMs) generate text by iteratively denoising token blocks,
exposing block-level parallelism that autoregressive (AR) decoding cannot exploit
~\cite{nie2025large,ye2025dream,wu2025fast2,arriola2025block,fu2025efficient}. The paradigm has scaled from
small-scale block-diffusion prototypes~\cite{arriola2025block}, through dense models~\cite{nie2025large,ye2025dream,wu2025fast2,fu2026nemotronlabsdiffusion}, 
to larger sparse
Mixture-of-Experts (MoE) models~\cite{zhu2025llada-moe,bie2025llada2-0,bie2026llada2-1}. 
Despite this block-level parallelism, DLLM decoding remains costly because high-quality generation requires repeatedly evaluating the target model to refine the active denoising block.
Even after optimizations such as block-wise decoding, confidence-based token revealing, adaptive caching,
dynamic cache eviction, and layer skipping~\cite{liu2025dllm,ma2025dkv,song2026sparse,goel2026skip,wu2025fast2,sun2026spa,wu2026blockbatch},
these repeated target-model evaluations remain the dominant cost~\cite{gao2025self,wu2025fast2}.

Speculative decoding offers a complementary route to reducing the number of denoising forward passes.
In AR models, speculative decoding accelerates generation by drafting candidate future
tokens and verifying them with the target model~\cite{leviathan2023fast,chen2023accelerating,cai2024medusa,li2024eagle}.
For latency-sensitive generation, \emph{multi-branch} variants further organize candidate continuations into draft graphs and verify multiple branches in parallel, trading additional parallel computation for fewer sequential decoding steps.
Recent work adapts this principle to DLLMs by constructing draft branch states and verifying them
together with the main branch in a single batched forward pass~\cite{agrawal2025spiffy,li2025diffuspec,gao2025self}.
This adaptation is structurally different from AR speculation. First, DLLM attention is bidirectional, so each draft is a full block state rather than a next-token continuation. Second, draft branches across denoising steps form a directed graph of related masked-token configurations. Existing \emph{multi-branch} speculative DLLM decoding methods verify one main branch and $B$ draft branches in one batched forward pass. If a draft matches the subsequent main-branch denoising trajectory, its precomputed logits allow skipping one or more future denoising forward passes.
Concretely, the decoding cost can be decomposed as
\begin{equation}
    \underbrace{\text{Total decoding cost}}_{\text{End-to-end}}
    \;=\;
    \underbrace{\text{number of denoising forward passes}}_{\costone \;  \text{Speculation}}
    \times
    \underbrace{\text{cost per forward pass}}_{\costtwo \; \text{Multi-branch execution}}.
    \label{eq:intro_two_levers}
\end{equation}

While multi-branch speculative DLLM decoding reduces \costone \; in Eq.~\ref{eq:intro_two_levers}, it also introduces a new bottleneck: each forward pass densely evaluates the main branch together with all draft branches, increasing cost per forward pass \costtwo\ as the branch count grows. The key observation of this paper is that this dense forward pass contains substantial \emph{multi-branch redundancy}. By construction, a draft inherits almost all tokens from its parent and differs only at
a small set of speculative unmasked positions. Consequently, most token representations in the child draft remain nearly identical to the corresponding parent representations across Transformer layers.
Our profiling on representative block-diffusion DLLMs demonstrates that more than $72\%$ of
measured draft--layer--position--step tuples have high parent--child hidden-state similarity, with divergence concentrated near recently unmasked positions. However, existing multi-branch speculative DLLM decoding methods process every draft branch densely, recomputing layers for representations that are already available from the parent draft
~\cite{agrawal2025spiffy,li2025diffuspec,gao2025self}.

To exploit this redundancy, we propose \textsc{SpecFold}, an algorithm--system co-design for
efficient multi-branch speculative DLLM decoding, reducing \costtwo \; in Eq.~\ref{eq:intro_two_levers}, the cost of each cross-branch execution call in multi-branch speculative DLLM decoding. At the algorithm level, \textsc{SpecFold}  builds a dynamic computation reuse relation across branches via a token-level parent--child residual gate. Similar positions reuse parent computation, which we refer to as \textit{folding} in the following text, while divergent positions are recomputed. Because the bi-directional attention in DLLM couples all positions, \textsc{SpecFold} reuses the parent's streaming attention state and corrects it with child's recomputed queries, keys and values. FFN outputs are reused at folded token positions, and each draft maintains its residual streams to preserve and accumulate cross-branch diversity.
At the system level, a Triton kernel implementation compacts recomputed token positions across branches, resolves fold chains, performs projection and FFN computation with packed recomputed positions, and applies the attention correction without materializing redundant branch tensors. Together, the algorithm exposes fine-grained cross-branch reuse and the system converts it into lower per-forward cost \costtwo \; in Eq.~\ref{eq:intro_two_levers}.

\textsc{SpecFold} is designed to be orthogonal to prior DLLM acceleration methods. Temporal caching~\cite{wu2025fast2, liu2025dllm,ma2025dkv} and layer skipping
reuse computation across denoising steps along a single trajectory
~\cite{goel2026skip}, whereas speculative
DLLM methods reduce the number of denoising forward passes by verifying drafts in parallel
~\cite{agrawal2025spiffy,li2025diffuspec,gao2025self}. \textsc{SpecFold} removes redundant computation \emph{within} each speculative verification step, across the drafts that are already evaluated together. It composes with existing speculation approaches such as Spiffy~\cite{agrawal2025spiffy} without modifying draft construction policies.
Our contributions are three-fold:

\begin{itemize}[leftmargin=1.4em,itemsep=0.1em,topsep=0.2em]
    \item We identify \emph{multi-branch activation redundancy} as a new efficiency opportunity in
    speculative DLLM decoding. Unlike temporal redundancy across denoising steps, this redundancy arises within a single speculative verification step.

    \item We introduce \textsc{SpecFold}, an algorithm--system co-design for folded multi-branch speculative DLLM decoding. At the algorithm level, \textsc{SpecFold} combines token-level residual gating, layer-wise hidden-state reuse, folded FFN, and attention correction, while maintaining branch hidden-state diversity. At the system level, a Triton kernel implementation maps these reuse decisions to sparse GPU execution that avoids redundant branch computation. The design is agnostic to the underlying draft-construction policy, orthogonal to temporal caching, and compatible with existing DLLM speculation strategies.

    \item 
    We show that \textsc{SpecFold} achieves up to $1.64\times$ throughput over Spiffy and $1.99\times$ over vanilla decoding across two DLLM families (Fast-dLLM-v2, Nemotron-Labs-Diffusion), five models, and five benchmarks, without compromising task performance.
\end{itemize}
\section{Background}
\label{sec:background}

\subsection{Masked Diffusion LLMs and Temporal Acceleration}
\label{sec:background:dllm}

Masked diffusion large language models (DLLMs) generate text by reversing a masking process
over token blocks~\cite{nie2025large,ye2025dream,arriola2025block}. During training, the model
observes a partially masked sequence and learns to reconstruct the original tokens. During inference,
generation proceeds block by block: for the current active block, the decoder starts from masked
positions and repeatedly predicts token distributions, unmasks a subset of positions, and refines the
remaining masked positions until the block is complete, as illustrated in Fig.~\ref{fig:pipeline}-(a). Unlike autoregressive (AR) decoding, a single
DLLM denoising step can update multiple token positions in parallel, exposing block-level parallelism.

We write \(k\) for the active block index and \(t\) for the denoising timestep within that block. Let
\(N\) denote the active-block length, and let \(X_k^{t} \in \mathcal{V}^{N}\) denote the current block
state, whose entries are either vocabulary tokens or the mask token \texttt{[M]}. The completed prefix,
prompt, and previously finalized blocks are treated as context \(C_k\). A DLLM forward pass can be
viewed as a map
\begin{equation}
    F_\theta(C_k, X_k^{t}) \rightarrow p_\theta(\cdot \mid C_k, X_k^{t}),
\end{equation}
which returns token distributions for the masked positions in the active block. An unmasking rule then
selects one or more high-confidence positions to unmask, producing the next state \(X_k^{t-1}\).

This iterative refinement is the main source of DLLM inference cost. Recent DLLM systems reduce
this cost by improving the denoising schedule or by reusing computation across adjacent denoising
states. These include block-wise decoding, confidence-based token revealing, adaptive activation
caching, dynamic cache eviction, and layer skipping~\cite{liu2025dllm,ma2025dkv,song2026sparse,goel2026skip,wu2025fast2,sun2026spa}.
Such methods exploit \emph{temporal redundancy}: adjacent states \(X_k^{t}\) and \(X_k^{t-1}\)
usually differ at only a few newly unmasked positions, so parts of the computation along a single
denoising trajectory can be reused. Multi-branch speculative DLLM decoding instead attempts to skip future
denoising forward passes entirely.

\subsection{Multi-Branch Speculative Decoding in DLLMs}
\label{sec:background:speculative}

Multi-branch speculative decoding accelerates generation by proposing candidate future states and verifying them
with the target model. In autoregressive language models, drafts are causal next-token continuations
that are accepted or rejected by the target model~\cite{leviathan2023fast,chen2023accelerating,cai2024medusa,li2024eagle}. For latency-sensitive generation, \emph{multi-branch} variants further organize candidate continuations into draft graphs and verify multiple branches in parallel, trading additional parallel computation for fewer sequential decoding steps.
Adapting this idea to DLLMs requires a different structure because DLLM attention is bidirectional. In DLLM, a draft is a full active-block state with a particular set of unmasked and masked positions.

At a multi-branch speculative DLLM verification step (see Fig.~\ref{fig:pipeline}-(b)), the decoder constructs a batch of \(B+1\) branch states:
one main branch \(X_{0,k}^{t}\) and \(B\) draft branches
\(\{X_{i,k}^{t}\}_{i=1}^{B}\). The branches form a directed draft graph rooted at the main branch. Each draft branch \(i\) has a
parent draft \(\operatorname{par}(i) \in \{0,\ldots,i-1\}\), so every branch's parent chain traces back to the root \(X_{0,k}^{t}\); it inherits the parent draft state and additionally unmasks at most \(S_t\) positions. Then, the target DLLM verifies all branches in a single batched forward pass:
\begin{equation}
\label{eq:dllm_spec_forward}
    p_{i,k}^{t} = F_\theta(C_k, X_{i,k}^{t}),
    \qquad i = 0,1,\ldots,B .
\end{equation}
The resulting draft--logit pairs \(\{(X_{i,k}^{t}, p_{i,k}^{t})\}_{i=1}^{B}\) are stored in a cache. If a
later main-branch state matches one of the cached draft states, the decoder can reuse the corresponding
precomputed logits and skip an additional denoising forward pass. In lossless speculative DLLM schemes, this
cache reuse preserves the target-model trajectory because the reused logits are
those produced by the same target model on the same block state~\cite{agrawal2025spiffy,li2025diffuspec,gao2025self}.

Existing multi-branch speculative DLLM methods differ in how they construct drafts and define acceptance, but commonly execute verification as a dense batch that computes every position of every branch~\cite{agrawal2025spiffy,li2025diffuspec,gao2025self}; we refer to this execution strategy as \emph{dense verification}. This reduces \costone\ in Eq.~\ref{eq:intro_two_levers}, but increases \costtwo\ by evaluating all branches densely within each forward pass. 

\subsection{Redundancy Exploitation in Transformer Inference}

Redundancy in Transformer inference has been reduced along two axes.
The first is the \textit{temporal axis}: reusing activations across
successive decoding or denoising steps, as in autoregressive KV caching, cross-request
prefix sharing \citep{kwon2023pagedattention,zheng2024sglang,
juravsky2024hydragen}, and cross-timestep feature caching in diffusion
models \citep{ma2024deepcache,ma2025dkv}.
The second is \textit{within-pass sparsity}: skipping or merging redundant tokens and layers inside a single forward pass via token merging and pruning \citep{bolya2023tome,wang2021spatten,rao2021dynamicvit} or adaptive early exit \citep{xin2020deebert,schuster2022calm}.
Multi-branch speculative decoding \citep{miao2024specinfer,cai2024medusa,
li2024eagle} batches a draft graph of continuations into one batched forward
pass and reuses the shared-prefix KV cache across graph nodes, but
recomputes every sibling branch's suffix activations independently.
\textsc{SpecFold} targets an unexploited third axis: \textit{multi-branch redundancy}
within each DLLM verification step for speculative drafts, folding the computation of draft branches, without
modifying the draft-construction policy or model weights.
\section{The Proposed \textsc{SpecFold} Framework}
\label{sec:sparspec}

\begin{figure*}[!t]
\centering
\includegraphics[width=\linewidth]{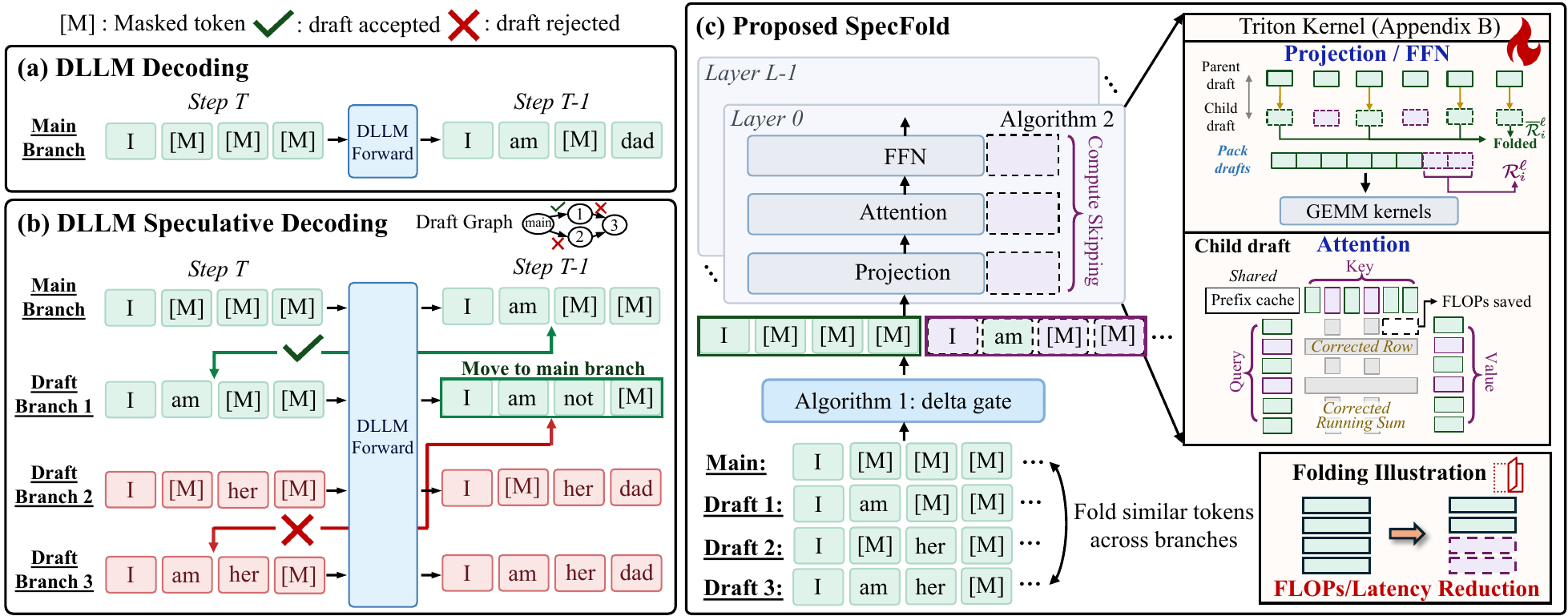}

\vspace{-0.8em}

\caption{
Overview of \textsc{SpecFold}.
(a) Standard DLLM decoding iteratively unmasks tokens through repeated denoising forward passes.
(b) Multi-branch speculative DLLM decoding verifies a main branch together with several draft branches in one batched forward pass. If a draft matches a subsequent main-branch state, its precomputed logits are reused to skip a future denoising forward pass.
(c) \textsc{SpecFold} exploits cross-branch redundancy within each verification step. Before every Transformer layer, a parent--child residual gate identifies folded positions, which reuse parent computation, while divergent positions are recomputed through the folded attention and FFN execution, improving decoding throughput.
}

\label{fig:pipeline}
\vspace{-1.0em}
\end{figure*}

\textsc{SpecFold} builds on the algorithmic observation that multi-branch speculative DLLM decoding exhibits strong cross-branch redundancy that dense computation fails to exploit.
As noted in Sec.~\ref{sec:background:speculative}, each draft branch inherits all but up to $S_t$ newly unmasked positions from its parent, suggesting that much of the per-branch computation in Eq.~\ref{eq:dllm_spec_forward} is redundant. We characterize this redundancy (Sec.~\ref{sec:sparspec:profile}), present the \textsc{SpecFold} algorithm (Sec.~\ref{sec:sparspec:strategies}), and describe the Triton kernel implementation that translates reused computation into throughput gains (Sec.~\ref{sec:sparspec:kernel}).

For notations, we write $\mathbf{H}^{\ell}_{i,k} \in \mathbb{R}^{N \times d}$ for the hidden state at input of Transformer layer $\ell$, denoising block $k$, and branch $i$, with $\mathbf{H}^{\ell}_{i,k}[n]$ denoting token position $n$. A per-layer gate on the parent-child hidden state residual magnitude determines folding for both the attention and FFN layers.
Fig.~\ref{fig:pipeline}-(c) summarizes \textsc{SpecFold}. Within a Transformer layer, token-level residual gating identifies folded token positions, parent branch computation is reused for projection and FFN, attention is corrected for child-specific updates, and the algorithmic sparsity is mapped to Triton kernels for higher decoding throughput.

\subsection{\textsc{SpecFold}: Profiling the Redundancy}
\label{sec:sparspec:profile}

\paragraph{Where redundancy lives.} By construction, each draft branch $X_{i,k}$ inherits all but up to $S_t$ newly unmasked positions from its parent $X_{\mathrm{par}(i),k}$. Therefore, we expect the corresponding layer-input hidden states to satisfy $\mathbf{H}^\ell_{i, k}[n] \approx \mathbf{H}^\ell_{\mathrm{par}(i), k}[n]$ on inherited positions $n$ at every layer $\ell$. That is, the computation for branch $X_{i,k}$ within the batched forward pass repeats compute that was already done for the parent, $X_{\mathrm{par}(i),k}$.

\begin{figure*}[t]
    \centering
    \includegraphics[
        width=1.0\textwidth,
        height=0.9\textheight,
        keepaspectratio
    ]{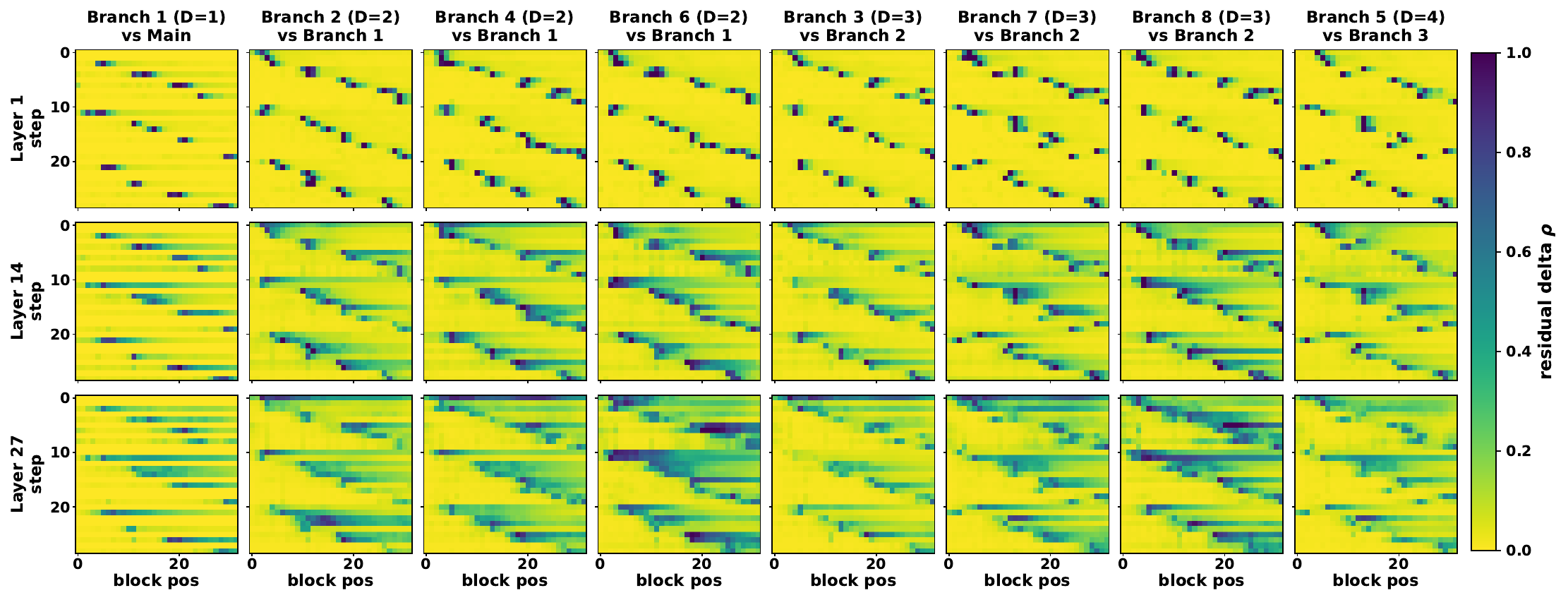}
    \vspace{-2.2em}
    \caption{
    Per-token relative residual $\rho$ of the layer-input hidden state between each draft branch and its immediate parent branch in the draft graph, recorded during the verification steps of Fast-dLLM-v2-7B decoding a GSM8K sample. Columns are the 8 draft branches ($D$ denotes a branch's depth in the draft graph), and rows are layers spanning the model depth. Within a panel, each row is one denoising step and each column one block position. The color indicates the level of divergence.
    }
    \vspace{-0.8em}
    \label{fig:similarity}
\end{figure*}

\paragraph{Empirical profile.} We test these predictions on redundancy during real verification steps of Fast-dLLM-v2-7B (block size $N\!=\!32$, calibrated draft graph with $B\!=\!8$ drafts). Fig.~\ref{fig:similarity} visualizes the relative residual $\rho^\ell_{i,k,n}$ between $\mathbf{H}^\ell_{i, k}[n]$ and $\mathbf{H}^\ell_{\mathrm{par}(i), k}[n]$ (defined in Eq.~\ref{eq:s2_sim}) across drafts ($i$), layers ($\ell$), denoising steps, and token positions ($n$) within the block. We conclude the following observations:

\textbf{Observation\,1: Hidden states match across parent--child pairs.}
Across all 28 layers, $72\%$ of (draft, step, layer, position) cells satisfy $\rho \leq 0.1$ and $88\%$ satisfy $\rho \leq 0.3$. High-divergence regions concentrate along the unmasking positions and the positions draft $i$ unmasked, where the child draft contributes new information. This supports that $\mathbf{H}^\ell_{i,k}[n]\!\approx\!\mathbf{H}^\ell_{\mathrm{par}(i),k}[n]$ on inherited positions $n$.

\begin{figure}[t]
    \centering
    \includegraphics[width=\linewidth]{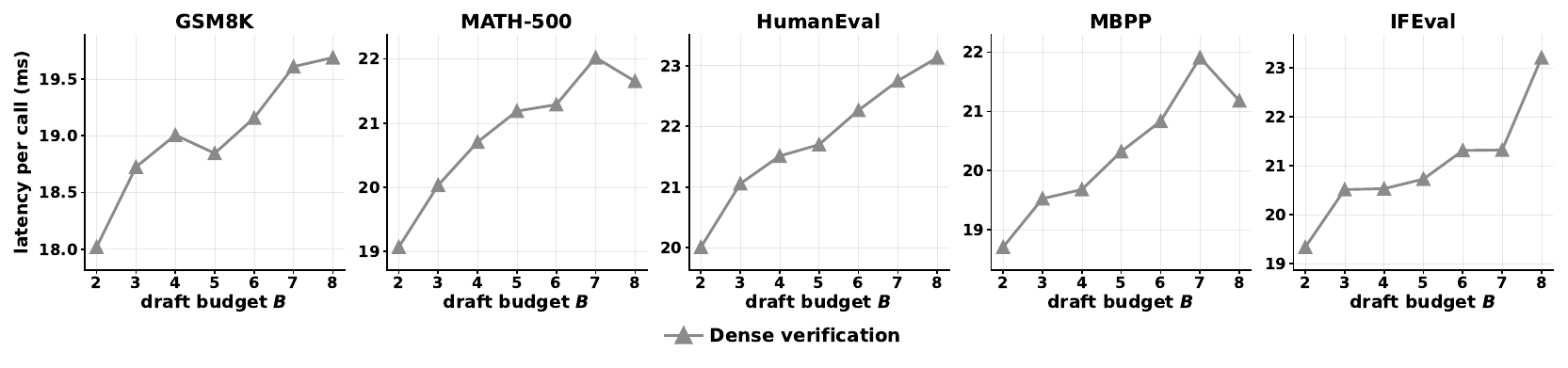}
    \vspace{-2.5em}
    \caption{
    Mean latency per forward pass of dense verification versus draft budget $B$ on
    Nemotron-Labs-Diffusion-3B on H100.
    }
    \vspace{-0.8em}
    \label{fig:percall}
\end{figure}

\textbf{Observation\,2: Dense verification pushes diffusion iteration to a compute-heavy workload.} We implemented Spiffy~\cite{agrawal2025spiffy} as a dense-verification baseline for DLLM decoding. Fig.~\ref{fig:percall} plots the mean latency per forward pass on Nemotron-Labs-Diffusion-3B as the draft budget grows from $B{=}2$ to $B{=}8$. Although the weights stream once for the whole batch, per-forward latency rises with every added draft branch across the five benchmarks, demonstrating that dense verification computes every position of every draft, paying latency per forward pass.

\subsection{\textsc{SpecFold}: Folding Redundant Drafts}
\label{sec:sparspec:strategies}

\textsc{SpecFold} exploits the cross-branch redundancy characterized in Sec.~\ref{sec:sparspec:profile}. As stated in Observation 1, for a parent--child draft pair $({\mathrm{par}(i)},\, i)$ among the speculative drafts, the hidden states $\mathbf{H}^\ell_{i,k}[n]$ at inherited token positions are nearly identical across all layers. Rather than recomputing these redundant activations, \textsc{SpecFold} identifies them via a per-token parent--child relative-residual gate before the QKV projection at each Transformer layer, and folds the child's computation into the parent's at those positions.

Specifically, at each layer $\ell$ and denoising block $k$, for each draft branch $i$ and each token position $n$, we compute the relative residual between $\mathbf{H}^\ell_{i,k}[n]$ and  $\mathbf{H}^\ell_{\mathrm{par}(i),k}[n]$:

\begin{equation}
    \rho^\ell_{i,k,n} \;=\;
    \frac{
        \bigl\| \mathbf{H}^\ell_{i,k}[n] - \mathbf{H}^\ell_{\mathrm{par}(i),k}[n] \bigr\|
    }{
        \bigl\| \mathbf{H}^\ell_{\mathrm{par}(i),k}[n] \bigr\|
    }.
    \label{eq:s2_sim}
\end{equation}

Given a recompute threshold $\delta$, if $\rho^\ell_{i,k,n} \leq \delta$, token $n$ of draft $i$ is \emph{folded} at layer $\ell$: its query, key, and value are inherited from the parent, and its attention and FFN outputs are obtained via the folded computation, detailed in Alg.~\ref{alg:specfoldlayer}. The resulting folded ($\mathcal{R}^\ell_i$) and recomputed ($\overline{\mathcal{R}}^\ell_i$) position sets are:
\begin{equation}
    \mathcal{R}^\ell_i \;=\; \bigl\{n : \rho^\ell_{i,k,n} \leq \delta\bigr\},
    \; \; \;
    \overline{\mathcal{R}}^\ell_i \;=\; [N] \setminus \mathcal{R}^\ell_i.
    \label{eq:specfold_sets}
\end{equation}

Alg.~\ref{alg:sparspec} summarizes one complete \textsc{SpecFold} step. We highlight a few design choices that materialize the throughput gain from the exploitation of cross-branch redundancy, while maintaining the hidden state diversity across speculative draft branches:

\begin{itemize}[leftmargin=1.2em,itemsep=1pt,topsep=1pt]
    \item \textbf{Maintaining diversity through residual path.} Across Transformer layers, each position in each draft branch maintains its residual stream since the residual add costs little computation. This design allows the hidden state diversity across branches to be propagated across layers.

    \item \textbf{Maintaining diversity through cross-position attention.} At self-attention layers, the attention output at a folded query position remains exact with respect to the branch's own KV cache by correcting the parent's streaming attention state, efficiently implemented with details in Sec.~\ref{sec:sparspec:kernel} and Appendix~\ref{appendix:foldedattn}. Across layers, each position's residual stream keeps accumulating its own branch's cross-position attention contributions through the residual add. Once these contributions push the divergence measure, $\rho^\ell_{i,k,n}$ above $\delta$, position $n$ recomputes, preserving branch diversity.

    \item \textbf{Cutting redundancy at projections and FFN layers.} The QKV projection and FFN layers are position-wise, so folded positions inherit the parent's outputs. These layers only run with the $|\overline{\mathcal{R}}^\ell_i|$ recomputed positions.
\end{itemize}

In Alg.~\ref{alg:sparspec}, there is a \textsc{SpecFoldLayer} procedure that materializes the computational reduction. See Appendix~\ref{appendix:foldedtransformer} for the detailed implementation.

\begin{algorithm}[t]
\footnotesize
\caption{One \textsc{SpecFold} step.}
\label{alg:sparspec}
\begin{algorithmic}[1]
\Require Branches $\{X^t_{i,k}\}_{i=0}^{B}$, recompute threshold $\delta$,
         prefix KV cache $\mathcal{C}$, LM head $W_{\mathrm{LM}}$, number of layers $L$
\Ensure  Output logits $\{p^t_{i,k}\}_{i=0}^{B}$ for all branches $i$
\State Embed all branches: $\mathbf{H}^{0}_{i,k} \leftarrow \mathrm{Embed}(X^t_{i,k})$
       for all $i \in \{0,\ldots,B\}$
\For{each layer $\ell = 0, \ldots, L-1$}
    \State $\overline{\mathcal{R}}^\ell_0 \leftarrow [N]$
           \hfill\Comment{main branch always recomputes}
    \For{each draft $i = 1, \ldots, B$}
        \State $\overline{\mathcal{R}}^\ell_i \leftarrow
               \{\, n : \|\mathbf{H}^{\ell}_{i,k}[n] - \mathbf{H}^{\ell}_{\mathrm{par}(i),k}[n]\|
               > \delta\, \|\mathbf{H}^{\ell}_{\mathrm{par}(i),k}[n]\| \,\}$
               \hfill\Comment{gate on hidden-state residuals}
        \State $\mathcal{R}^\ell_i \leftarrow [N] \setminus \overline{\mathcal{R}}^\ell_i$
               \hfill\Comment{folded / recompute sets}
    \EndFor
    \State $\{\mathbf{H}^{\ell+1}_{i,k}\}_{i=0}^{B} \leftarrow
           \Call{SpecFoldLayer}{\{\mathbf{H}^\ell_{i,k}\}_{i=0}^{B},\,
                \mathrm{par}(\cdot),\, \{\overline{\mathcal{R}}^\ell_i\}_{i=0}^{B},\,
                \mathcal{C}_\ell}$
           \hfill\Comment{Alg.~\ref{alg:specfoldlayer}}
\EndFor
\State $p^t_{i,k} \leftarrow W_{\mathrm{LM}}\,\mathrm{Norm}(\mathbf{H}^{L}_{i,k})$
       for all $i \in \{0,\ldots,B\}$
\State Cache $\{(X^t_{i,k},\, p^t_{i,k})\}_{i=0}^{B}$ for speculative verification
\end{algorithmic}
\end{algorithm}

\subsection{Triton Implementation}
\label{sec:sparspec:kernel}

To materialize the algorithmic computation reduction into actual end-to-end throughput gains, we implement each \textsc{SpecFoldLayer} as a set of efficient Triton kernels. For gating in Eq.~\ref{eq:s2_sim}, a compact kernel evaluates the residual, records the source of folded positions, and compacts all recomputed positions across draft branches. This packing allows QKV projections and FFNs to operate on this packed batch. A source-resolution kernel walks the fold chain, enabling every folded position to access the parent state.

For attention layers, we implement a FlashAttention-style streaming kernel~\cite{dao2022flashattention}. For recomputed positions, query vectors execute streaming attention over the shared prefix and each branch's KV cache. For folded positions, query vectors reuse the streaming attention state of the parent and correct it for recomputed keys and values within the branch. The shared prefix KV cache is expanded across branches as a zero-copy strided view.

This implementation supports both evaluated model families, including grouped-query attention. Appendix~\ref{appendix:triton} details the algorithm and kernel implementation.
\newcommand{\ginc}[1]{\,\textcolor{OliveGreen}{\scriptsize($\uparrow$#1)}}

\section{Experiments}
\label{sec:eval}

\subsection{Setup}

\textbf{Models and baselines.}
We evaluate two DLLM families: Fast-dLLM-v2~\cite{wu2025fast2} (1.5B, 7B) and Nemotron-Labs-Diffusion (NLD)~\cite{fu2026nemotronlabsdiffusion} (3B, 8B, 14B). For Fast-dLLM-v2, we follow the default configuration with block size $32$, small-block size $8$, confidence threshold $0.9$, and block caching disabled. For Nemotron-Labs-Diffusion, we use block size $32$, confidence threshold $0.9$, and refresh the prefix KV cache with a causal forward pass at each block boundary. Across all models, Spiffy constructs the draft graph and \textsc{SpecFold} reuses the same graph for fair comparison.
We compare the following decoding strategies:
\begin{itemize}[leftmargin=1.2em,itemsep=0pt,topsep=0.2pt]
    \item \emph{Vanilla decoding.}
    The native block-wise decoding of each model, without speculative drafts.

    \item \emph{Spiffy}~\cite{agrawal2025spiffy}.
    A calibration-based DLLM speculative decoding baseline. Following Spiffy, we calibrate one degree-1 draft graph per model on $50$ samples, and instantiate $B{=}7$ draft branches at inference time according to the calibrated token-position and vocabulary-rank patterns. The calibration samples contains $25$ from the MBPP training split and $25$ from the MATH training split, disjoint from all evaluation data. For each model, a draft graph serves every benchmark.

    \item \emph{\textsc{SpecFold}.}
    \textsc{SpecFold} is our proposed method with folded cross-branch computation, where we use the same draft graph construction algorithm as Spiffy.
\end{itemize}

\textbf{Benchmarks, Metrics, and Hardware.}
Our evaluation spans three domains: mathematical reasoning (GSM8K~\cite{cobbe2021training} and MATH~\cite{hendrycks2021measuring}), code generation (HumanEval~\cite{chen2021evaluating} and MBPP~\cite{austin2021program}), and instruction following (IFEval~\cite{zhou2023instruction}).
Following Fast-dLLM-v2's evaluation setup, we use zero-shot prompting for all tasks.
Non-code benchmarks are evaluated with LM-Eval, and HumanEval and MBPP are evaluated with EvalPlus~\cite{liu2023is}.
We report three primary metrics: Number of Function Evaluations (NFE) per sample, tokens per second (TPS), and accuracy.
NFE measures the number of denoising forward passes required for generation, while TPS measures end-to-end decoding speed on hardware.
All experiments, unless otherwise stated, are run on a single NVIDIA H100 80GB GPU.

% =========================
% Table: All five models on H100
% =========================
\begin{table*}[t]
\centering
\caption{\textsc{SpecFold} evaluation, reporting per-sample NFE, decoding
throughput (tokens/s), and accuracy across five benchmarks and five model
scales spanning the Fast-dLLM-v2 and Nemotron-Labs-Diffusion families.
\textsc{SpecFold} uses recompute threshold $\delta=0.1$. Nemotron-Labs-Diffusion models run in diffusion mode. The MATH row uses the MATH-500 test set.}
\label{tab:h100_all}
\vspace{-1em}
\small
\setlength{\tabcolsep}{3pt}
\resizebox{\textwidth}{!}{%
\begin{tabular}{llccc|ccc|ccc|ccc|ccc}
\toprule
\multirow{2}{*}{\textbf{Benchmark}}
& \multirow{2}{*}{\textbf{Method}}
& \multicolumn{3}{c|}{\textbf{Fast-dLLM-v2-1.5B}}
& \multicolumn{3}{c|}{\textbf{Fast-dLLM-v2-7B}}
& \multicolumn{3}{c|}{\textbf{NLD-3B}}
& \multicolumn{3}{c|}{\textbf{NLD-8B}}
& \multicolumn{3}{c}{\textbf{NLD-14B}} \\
\cmidrule(lr){3-5}
\cmidrule(lr){6-8}
\cmidrule(lr){9-11}
\cmidrule(lr){12-14}
\cmidrule(lr){15-17}
&
& \textbf{NFE$\downarrow$} & \textbf{TPS$\uparrow$} & \textbf{Acc.$\uparrow$}
& \textbf{NFE$\downarrow$} & \textbf{TPS$\uparrow$} & \textbf{Acc.$\uparrow$}
& \textbf{NFE$\downarrow$} & \textbf{TPS$\uparrow$} & \textbf{Acc.$\uparrow$}
& \textbf{NFE$\downarrow$} & \textbf{TPS$\uparrow$} & \textbf{Acc.$\uparrow$}
& \textbf{NFE$\downarrow$} & \textbf{TPS$\uparrow$} & \textbf{Acc.$\uparrow$} \\
\midrule

\multirow{3}{*}{GSM8K}
& Vanilla
& 113 & 145 & 63.08\%
& 113 & 170 & 82.41\%
& 112 & 144 & 88.02\%
& 128 & 102 & 84.84\%
& 113 & 84 & 83.02\% \\
& Spiffy
& 90 & 150 & 62.77\%
& 95 & 173 & 83.09\%
& 86 & 161 & 87.95\%
& 93 & 117 & 84.84\%
& 82 & 98 & 83.47\% \\
\rowcolor{colOurs}
& \textbf{\textsc{SpecFold}}
& 90 & 229\ginc{58\%} & 62.55\%
& 96 & 188\ginc{11\%} & 81.96\%
& 86 & 204\ginc{42\%} & 88.17\%
& 93 & 131\ginc{29\%} & 84.61\%
& 82 & 102\ginc{21\%} & 82.87\% \\

\midrule
\multirow{3}{*}{MATH}
& Vanilla
& 263 & 173 & 38.60\%
& 229 & 160 & 57.80\%
& 225 & 156 & 72.60\%
& 297 & 113 & 83.80\%
& 246 & 104 & 84.00\% \\
& Spiffy
& 219 & 189 & 37.40\%
& 189 & 163 & 57.40\%
& 172 & 164 & 72.60\%
& 219 & 126 & 82.20\%
& 183 & 110 & 84.60\% \\
\rowcolor{colOurs}
& \textbf{\textsc{SpecFold}}
& 214 & 266\ginc{54\%} & 36.60\%
& 193 & 193\ginc{21\%} & 57.40\%
& 171 & 211\ginc{35\%} & 72.80\%
& 224 & 139\ginc{23\%} & 82.20\%
& 184 & 115\ginc{10\%} & 85.60\% \\

\midrule
\multirow{3}{*}{HumanEval}
& Vanilla
& 161 & 133 & 39.02\%
& 156 & 117 & 65.85\%
& 286 & 110 & 73.78\%
& 464 & 86 & 66.46\%
& 343 & 77 & 71.34\% \\
& Spiffy
& 122 & 147 & 40.24\%
& 117 & 135 & 64.63\%
& 212 & 115 & 75.61\%
& 314 & 99 & 68.29\%
& 237 & 87 & 71.34\% \\
\rowcolor{colOurs}
& \textbf{\textsc{SpecFold}}
& 114 & 209\ginc{58\%} & 40.24\%
& 108 & 142\ginc{21\%} & 61.59\%
& 209 & 154\ginc{40\%} & 70.12\%
& 308 & 111\ginc{28\%} & 68.90\%
& 232 & 92\ginc{20\%} & 74.39\% \\

\midrule
\multirow{3}{*}{MBPP}
& Vanilla
& 153 & 121 & 48.68\%
& 163 & 121 & 68.78\%
& 255 & 113 & 73.02\%
& 247 & 87 & 83.07\%
& 299 & 81 & 79.89\% \\
& Spiffy
& 118 & 148 & 48.41\%
& 114 & 133 & 67.99\%
& 191 & 121 & 70.37\%
& 165 & 103 & 83.86\%
& 206 & 91 & 80.69\% \\
\rowcolor{colOurs}
& \textbf{\textsc{SpecFold}}
& 121 & 242\ginc{99\%} & 49.21\%
& 120 & 155\ginc{28\%} & 68.52\%
& 190 & 158\ginc{40\%} & 72.75\%
& 170 & 114\ginc{32\%} & 83.33\%
& 207 & 95\ginc{18\%} & 79.63\% \\

\midrule
\multirow{3}{*}{IFEval}
& Vanilla
& 294 & 132 & 48.43\%
& 270 & 111 & 64.88\%
& 234 & 94 & 60.26\%
& 313 & 96 & 59.89\%
& 304 & 73 & 62.11\% \\
& Spiffy
& 226 & 140 & 47.50\%
& 200 & 119 & 64.33\%
& 159 & 106 & 60.63\%
& 215 & 105 & 58.41\%
& 211 & 84 & 61.18\% \\
\rowcolor{colOurs}
& \textbf{\textsc{SpecFold}}
& 225 & 222\ginc{68\%} & 45.84\%
& 204 & 145\ginc{31\%} & 65.06\%
& 167 & 139\ginc{48\%} & 58.96\%
& 213 & 124\ginc{29\%} & 61.55\%
& 209 & 90\ginc{24\%} & 60.81\% \\

\bottomrule
\end{tabular}
}
\vspace{-0.8em}
\end{table*}

\subsection{Main Results}

Table~\ref{tab:h100_all} presents our primary results across all five models.
Both efficiency metrics (NFE, TPS) are reported relative to Vanilla decoding: NFE reductions capture the denoising forward passes skipped by speculative verification, while TPS gains capture the end-to-end effect of \textsc{SpecFold} folding the compute of dense verification.

\paragraph{Fast-dLLM-v2.}
\textsc{SpecFold} reduces the cost of dense verification at both model scales. Across the five benchmarks, it achieves up to $1.64\times$ throughput over Spiffy and $1.99\times$ over Vanilla on Fast-dLLM-v2-1.5B, and up to $1.22\times$ over Spiffy and $1.31\times$ over Vanilla on Fast-dLLM-v2-7B. These results show that \textsc{SpecFold} converts the NFE savings of multi-branch speculative decoding into end-to-end speed by removing redundant computation introduced by dense verification while maintaining comparable accuracy. The accuracy drifts in Spiffy are floating-point noise, where we detailed this effect in Appendix~\ref{appendix:numerical_discrepancy} and justified the correctness of our implementation with a serialized-execution sanity check.

\paragraph{Nemotron-Labs-Diffusion.}
Across the 3B, 8B, and 14B models, \textsc{SpecFold} achieves up to $1.34\times$, $1.18\times$, and $1.07\times$ throughput over Spiffy, respectively, and up to $1.48\times$, $1.32\times$, and $1.24\times$ over Vanilla. The accuracy of \textsc{SpecFold} remains comparable to both Vanilla and Spiffy across all five benchmarks, with small deviations in both directions. Overall, the results show that \textsc{SpecFold} provides consistent end-to-end throughput gains across block-diffusion model families and scales without compromising accuracy.

\subsection{Ablations}
\label{sec:ablation}

\begin{figure}[t]
    \centering
    \includegraphics[width=\linewidth]{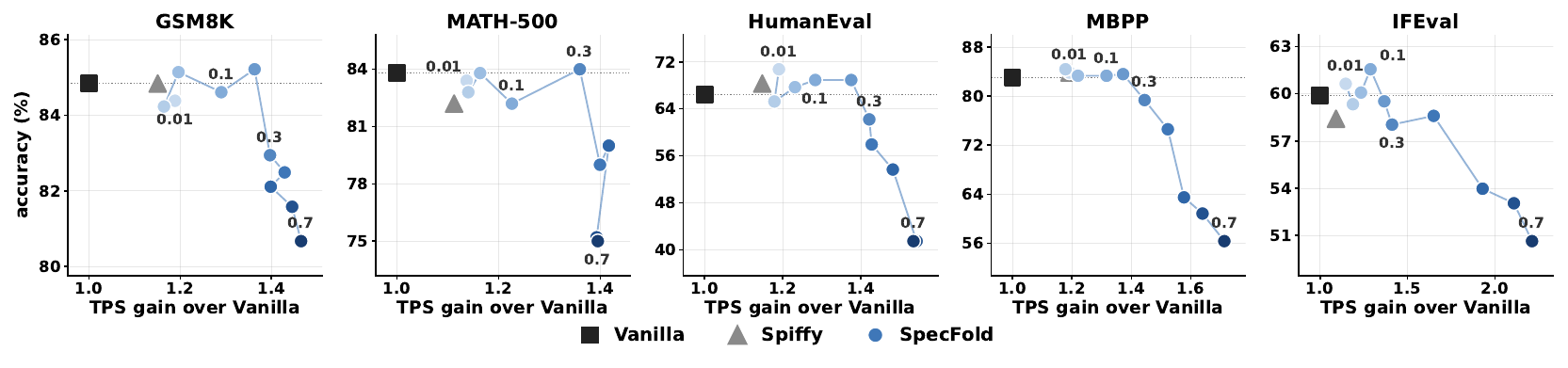}
    \vspace{-2.5em}
    \caption{
    Recompute-threshold sweep of \textsc{SpecFold} on Nemotron-Labs-Diffusion-8B (H100, $B{=}7$).
    Each panel plots task accuracy against the throughput (TPS) gain over Vanilla as $\delta$ increases from $0.01$ to $0.7$ (darker markers = larger $\delta$, i.e., more folding); the dotted line marks the Vanilla accuracy.
    Upper-right is better.
    }
    \vspace{-0.8em}
    \label{fig:delta_sweep_nld8b}
\end{figure}

\textbf{Recompute-threshold sweep.}
We evaluate the sensitivity of \textsc{SpecFold} to the recompute threshold $\delta$ on Nemotron-Labs-Diffusion-8B across all five benchmarks, sweeping
$\delta\in\{0.01, 0.02, 0.05, 0.1, 0.2, 0.3, 0.4, 0.5, 0.6, 0.7\}$
while keeping all other decoding hyperparameters fixed. Fig.~\ref{fig:delta_sweep_nld8b} plots the resulting accuracy--throughput trade-off per benchmark, with Vanilla and Spiffy shown as reference points.

\textsc{SpecFold} exhibits a broad and robust operating region in the latency-accuracy trade-off relation. Across all five benchmarks, accuracy stays at Vanilla's level through $\delta=0.2$ while folding already delivers a $1.3$--$1.4\times$ throughput gain (TPS) over Vanilla. Accuracy declines beyond this plateau. Measured throughput continues to grow with $\delta$, but gains beyond $\delta=0.3$ come at the cost of accuracy. Moreover, when $\delta$ approaches $0$, forcing all drafts to be recomputed, the accuracy--throughput point approaches Spiffy across all benchmarks, validating our implementation.

These results indicate that the efficiency gains do not depend on a precise choice of $\delta$. For the main evaluation in the paper, we fix $\delta=0.1$ as a conservative default across all models and benchmarks without per-task tuning. 

\begin{figure}[t]
    \centering
    \includegraphics[width=\linewidth]{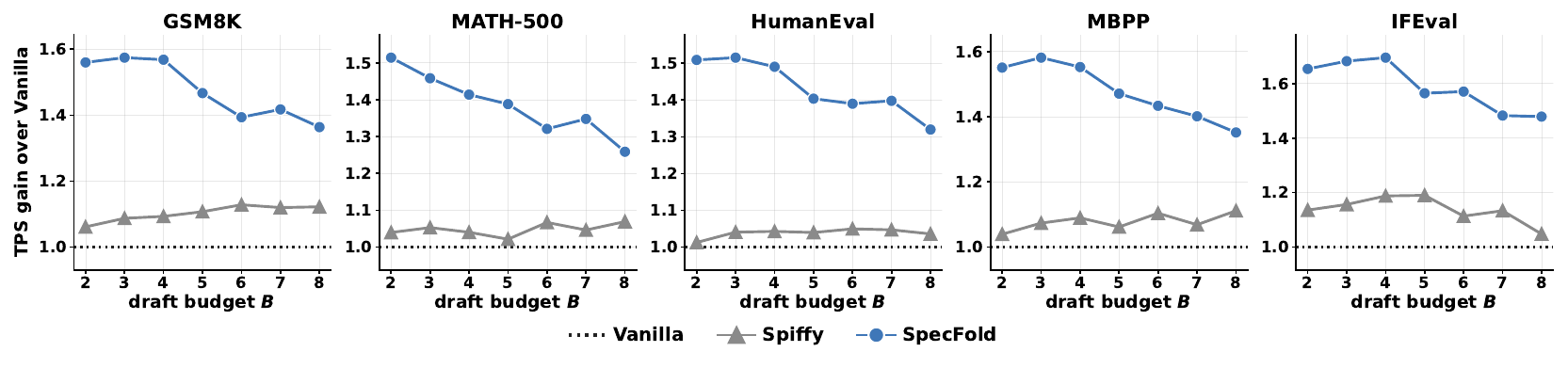}
    \vspace{-2.5em}
    \caption{
    Draft-budget sweep on Nemotron-Labs-Diffusion-3B, fixing $\delta{=}0.1$. Each panel plots the throughput gain (TPS) over Vanilla against the draft budget $B$. The gap between the two curves demonstrates the effect of folding.
    }
    \vspace{-0.8em}
    \label{fig:bscan_nld3b}
\end{figure}

\textbf{Draft-budget sweep.}
We analyze the sensitivity and effectiveness of \textsc{SpecFold} to the number of speculative draft branches $B$ on Nemotron-Labs-Diffusion-3B across all five benchmarks, sweeping $B \in \{2,\ldots,8\}$. Each budget uses its own calibrated draft graph with $B$ nodes. Fig.~\ref{fig:bscan_nld3b} plots the throughput gain (TPS) over Vanilla of Spiffy and \textsc{SpecFold} at every $B$.

Generally, the acceptance of speculative drafts grows with $B$ and saturates around $B=6$, where NFE reductions reach $30\%$ relative to Vanilla. Throughput, however, distinguishes the effectiveness of \textsc{SpecFold}. Spiffy stays within $0.99$--$1.17\times$ of Vanilla's TPS across the entire sweep due to the cost of dense verification. \textsc{SpecFold}, which shares the identical drafts and NFE at every $B$, converts the same acceptance into up to $1.73\times$ throughput gain. Appendix~\ref{app:bscan_all} repeats this sweep on the remaining four models and demonstrates the same structure across scales and families. Furthermore, Appendix~\ref{app:depth_ablation} varies the draft-graph depth at a fixed budget and shows that \textsc{SpecFold}'s advantage is consistent across drafting strategies. Appendix~\ref{app:gpu_ablation} validates \textsc{SpecFold}'s effectiveness across different GPUs.

\subsection{Analysis of the Residual Ripple}
\label{sec:observation}

\newcommand{\norm}[1]{\left\lVert#1\right\rVert}

\begin{figure}[t]
    \centering
    \includegraphics[width=1.0\linewidth]{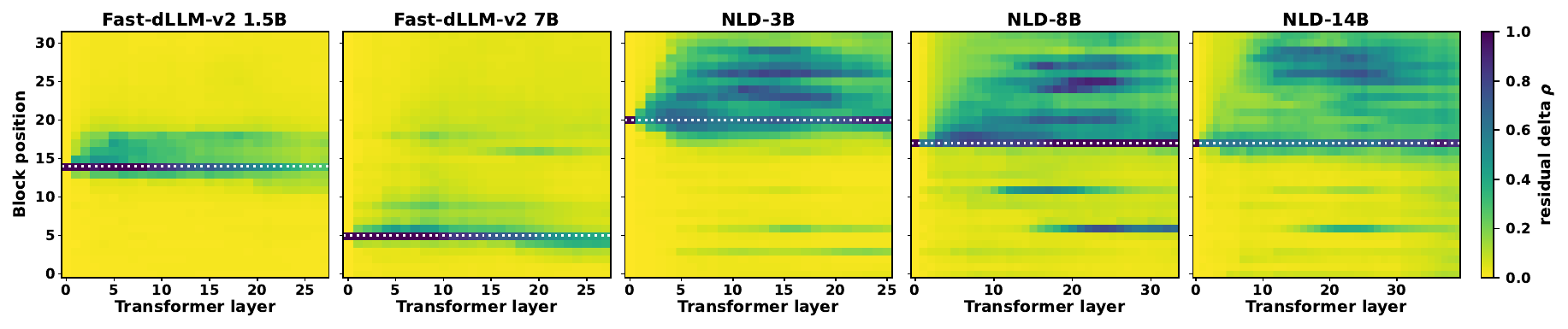}
    \vspace{-2em}
    \caption{\textbf{Per-position relative residual between a parent draft and a child draft that unmasks a single token.} Each model decodes a GSM8K prompt to the middle of a block: the parent draft is the block state with 17 of the 32 positions committed, and the child draft adds one more unmasked token being the target's next most-confident masked position (the ``edited position'', white dotted line). Heatmaps show the per-position parent--child relative residual $\rho^\ell_n$ (Eq.~\ref{eq:s2_sim}) across layers.}
    \label{fig:ripple}
\end{figure}

To test whether the layer-input hidden state $\mathbf{H}^\ell_{i,k}[n]$ provides a reliable basis for the folding decision in Eq.~\ref{eq:s2_sim}, we compare a parent-child $(\mathrm{par}(i), i)$ pair that differs by only one unmasked token in the speculative draft branch. Fig.~\ref{fig:ripple} shows the structure of resulting parent--child relative residual $\rho^\ell_n$.

The edited position creates a persistent high-residual divergence across layers. Interestingly, the residual divergence elsewhere splits by decoding state. Residual divergence $\rho$ stays low at committed positions, while the divergence ripple concentrates in the masked positions and recently unmasked positions. This observation motivates a per-token gate in Eq.~\ref{eq:s2_sim}, providing folding decisions in a higher granularity.
\section{Conclusion}
\label{sec:conclusion}

We propose \textsc{SpecFold}, an algorithm--system co-design for accelerating multi-branch speculative DLLM verification by exploiting \emph{cross-branch activation redundancy}. \textsc{SpecFold} combines token-level folding with a dedicated kernel implementation to reduce redundant computation within each speculative verification step. Across two DLLM families, five models, and five benchmarks, \textsc{SpecFold} achieves up to $1.64\times$ throughput over Spiffy and up to $1.99\times$ over Vanilla decoding while maintaining comparable task performance, unlocking the effective speedup with speculative DLLM decoding.

\subsection*{Ethics statement}

This paper presents system, empirical and methodological contributions and does not involve human subjects, data collection, or the release of new datasets or models. We do not foresee any ethical concerns arising from this work.

\bibliography{iclr2027_conference}

@inproceedings{leviathan2023fast,
  title        = {Fast Inference from Transformers via Speculative Decoding},
  author       = {Leviathan, Yaniv and Kalman, Matan and Matias, Yossi},
  booktitle    = {International Conference on Machine Learning},
  pages        = {19274--19286},
  year         = {2023},
  organization = {PMLR}
}

@misc{chen2023accelerating,
  title         = {Accelerating Large Language Model Decoding with Speculative Sampling},
  author        = {Chen, Charlie and Borgeaud, Sebastian and Irving, Geoffrey and Lespiau, Jean-Baptiste and Sifre, Laurent and Jumper, John},
  year          = {2023},
  eprint        = {2302.01318},
  archivePrefix = {arXiv},
  primaryClass  = {cs.CL}
}

@misc{cai2024medusa,
  title         = {{Medusa}: Simple {LLM} Inference Acceleration Framework with Multiple Decoding Heads},
  author        = {Cai, Tianle and Li, Yuhong and Geng, Zhengyang and Peng, Hongwu and Lee, Jason D. and Chen, Deming and Dao, Tri},
  year          = {2024},
  eprint        = {2401.10774},
  archivePrefix = {arXiv},
  primaryClass  = {cs.LG}
}

@misc{li2024eagle,
  title         = {{EAGLE}: Speculative Sampling Requires Rethinking Feature Uncertainty},
  author        = {Li, Yuhui and Wei, Fangyun and Zhang, Chao and Zhang, Hongyang},
  year          = {2024},
  eprint        = {2401.15077},
  archivePrefix = {arXiv},
  primaryClass  = {cs.CL}
}

@inproceedings{dao2022flashattention,
  title     = {{FlashAttention}: Fast and Memory-Efficient Exact Attention with {IO}-Awareness},
  author    = {Dao, Tri and Fu, Dan and Ermon, Stefano and Rudra, Atri and R{\'e}, Christopher},
  booktitle = {Advances in Neural Information Processing Systems},
  volume    = {35},
  pages     = {16344--16359},
  year      = {2022}
}

@article{goel2026skip,
  title         = {Skip to the Good Part: Representation Structure and Inference-Time Layer Skipping in Diffusion vs. Autoregressive {LLMs}},
  author        = {Goel, Raghavv and Garrepalli, Risheek and Agrawal, Sudhanshu and Lott, Chris and Lee, Mingu and Porikli, Fatih},
  journal       = {arXiv preprint arXiv:2603.07475},
  year          = {2026},
  eprint        = {2603.07475},
  archivePrefix = {arXiv}
}

@misc{sun2026spa,
  title         = {{SPA-Cache}: Singular Proxies for Adaptive Caching in Diffusion Language Models},
  author        = {Sun, Wenhao and Tu, Rong-Cheng and Ding, Yifu and Jin, Zhao and Liao, Jingyi and Jing, Yongcheng and Tao, Dacheng},
  year          = {2026},
  eprint        = {2602.02544},
  archivePrefix = {arXiv},
  primaryClass  = {cs.CL}
}

@misc{cobbe2021training,
  title         = {Training Verifiers to Solve Math Word Problems},
  author        = {Cobbe, Karl and Kosaraju, Vineet and Bavarian, Mohammad and Chen, Mark and Jun, Heewoo and Kaiser, Lukasz and Plappert, Matthias and Tworek, Jerry and Hilton, Jacob and Nakano, Reiichiro and Hesse, Christopher and Schulman, John},
  year          = {2021},
  eprint        = {2110.14168},
  archivePrefix = {arXiv},
  primaryClass  = {cs.LG}
}

@inproceedings{hendrycks2021measuring,
  title     = {Measuring Mathematical Problem Solving with the {MATH} Dataset},
  author    = {Hendrycks, Dan and Burns, Collin and Kadavath, Saurav and Arora, Akul and Basart, Steven and Tang, Eric and Song, Dawn and Steinhardt, Jacob},
  booktitle = {Advances in Neural Information Processing Systems},
  volume    = {34},
  year      = {2021}
}

@misc{chen2021evaluating,
  title         = {Evaluating Large Language Models Trained on Code},
  author        = {Chen, Mark and Tworek, Jerry and Jun, Heewoo and Yuan, Qiming and Pinto, Henrique Ponde de Oliveira and Kaplan, Jared and Edwards, Harri and Burda, Yuri and Joseph, Nicholas and Brockman, Greg and others},
  year          = {2021},
  eprint        = {2107.03374},
  archivePrefix = {arXiv},
  primaryClass  = {cs.LG}
}

@misc{zhou2023instruction,
  title         = {Instruction-Following Evaluation for Large Language Models},
  author        = {Zhou, Jeffrey and Lu, Tianjian and Mishra, Swaroop and Brahma, Siddhartha and Basu, Sujoy and Luan, Yi and Zhou, Denny and Hou, Le},
  year          = {2023},
  eprint        = {2311.07911},
  archivePrefix = {arXiv},
  primaryClass  = {cs.CL}
}

@inproceedings{liu2023is,
  title     = {Is Your Code Generated by {ChatGPT} Really Correct? Rigorous Evaluation of Large Language Models for Code Generation},
  author    = {Liu, Jiawei and Xia, Chunqiu Steven and Wang, Yuyao and Zhang, Lingming},
  booktitle = {Advances in Neural Information Processing Systems},
  year      = {2023},
  url       = {https://openreview.net/forum?id=1qvx610Cu7}
}

@article{wu2025fast2,
  title         = {{Fast-dLLM} v2: Efficient Block-Diffusion {LLM}},
  author        = {Wu, Chengyue and Zhang, Hao and Xue, Shuchen and Diao, Shizhe and Fu, Yonggan and Liu, Zhijian and Molchanov, Pavlo and Luo, Ping and Han, Song and Xie, Enze},
  journal       = {arXiv preprint arXiv:2509.26328},
  year          = {2025},
  eprint        = {2509.26328},
  archivePrefix = {arXiv}
}

@article{nie2025large,
  title         = {Large Language Diffusion Models},
  author        = {Nie, Shen and Zhu, Fengqi and You, Zebin and Zhang, Xiaolu and Ou, Jingyang and Hu, Jun and Zhou, Jun and Lin, Yankai and Wen, Ji-Rong and Li, Chongxuan},
  journal       = {arXiv preprint arXiv:2502.09992},
  year          = {2025},
  eprint        = {2502.09992},
  archivePrefix = {arXiv}
}

@article{ye2025dream,
  title         = {{Dream} 7B: Diffusion Large Language Models},
  author        = {Ye, Jiacheng and Xie, Zhihui and Zheng, Lin and Gao, Jiahui and Wu, Zirui and Jiang, Xin and Li, Zhenguo and Kong, Lingpeng},
  journal       = {arXiv preprint arXiv:2508.15487},
  year          = {2025},
  eprint        = {2508.15487},
  archivePrefix = {arXiv}
}

@misc{zhu2025llada-moe,
  title         = {{LLaDA-MoE}: A Sparse {MoE} Diffusion Language Model},
  author        = {Fengqi Zhu and Zebin You and Yipeng Xing and Zenan Huang and Lin Liu and Yihong Zhuang and Guoshan Lu and Kangyu Wang and Xudong Wang and Lanning Wei and Hongrui Guo and Jiaqi Hu and Wentao Ye and Tieyuan Chen and Chenchen Li and Chengfu Tang and Haibo Feng and Jun Hu and Jun Zhou and Xiaolu Zhang and Zhenzhong Lan and Junbo Zhao and Da Zheng and Chongxuan Li and Jianguo Li and Ji-Rong Wen},
  year          = {2025},
  eprint        = {2509.24389},
  archivePrefix = {arXiv},
  primaryClass  = {cs.CL},
  url           = {https://arxiv.org/abs/2509.24389}
}

@article{arriola2025block,
  title         = {Block Diffusion: Interpolating Between Autoregressive and Diffusion Language Models},
  author        = {Arriola, Marianne and Gokaslan, Aaron and Chiu, Justin T. and Yang, Zhihan and Qi, Zhixuan and Han, Jiaqi and Sahoo, Subham Sekhar and Kuleshov, Volodymyr},
  journal       = {arXiv preprint arXiv:2503.09573},
  year          = {2025},
  eprint        = {2503.09573},
  archivePrefix = {arXiv}
}

@inproceedings{
agrawal2025spiffy,
title={Structuring The Future: Diffusion {LLM} Speculative Decoding via Calibrated Draft Graphs},
author={Sudhanshu Agrawal and Risheek Garrepalli and Raghavv Goel and Christopher Lott and Fatih Porikli and Mingu Lee},
booktitle={ICML 2026 Workshop on Structured Probabilistic Inference {\&} Generative Modeling},
year={2026},
url={https://openreview.net/forum?id=LHwa2TCan6}
}

@article{li2025diffuspec,
  title         = {{DiffuSpec}: Unlocking Diffusion Language Models for Speculative Decoding},
  author        = {Li, Guanghao and Fu, Zhihui and Fang, Min and Zhao, Qibin and Tang, Ming and Yuan, Chun and Wang, Jun},
  journal       = {arXiv preprint arXiv:2510.02358},
  year          = {2025},
  eprint        = {2510.02358},
  archivePrefix = {arXiv}
}

@article{gao2025self,
  title         = {Self Speculative Decoding for Diffusion Large Language Models},
  author        = {Gao, Yifeng and Ji, Ziang and Wang, Yuxuan and Qi, Biqing and Xu, Hanlin and Zhang, Linfeng},
  journal       = {arXiv preprint arXiv:2510.04147},
  year          = {2025},
  eprint        = {2510.04147},
  archivePrefix = {arXiv}
}

@inproceedings{ma2024deepcache,
  title     = {{DeepCache}: Accelerating Diffusion Models for Free},
  author    = {Ma, Xinyin and Fang, Gongfan and Wang, Xinchao},
  booktitle = {Proceedings of the IEEE/CVF Conference on Computer Vision and Pattern Recognition},
  pages     = {15762--15772},
  year      = {2024}
}

@article{ma2025dkv,
  title         = {{dKV-Cache}: The Cache for Diffusion Language Models},
  author        = {Ma, Xinyin and Yu, Runpeng and Fang, Gongfan and Wang, Xinchao},
  journal       = {arXiv preprint arXiv:2505.15781},
  year          = {2025},
  eprint        = {2505.15781},
  archivePrefix = {arXiv}
}

@article{liu2025dllm,
  title         = {{dLLM-Cache}: Accelerating Diffusion Large Language Models with Adaptive Caching},
  author        = {Liu, Zhiyuan and Yang, Yicun and Zhang, Yaojie and Chen, Junjie and Zou, Chang and Wei, Qingyuan and Wang, Shaobo and Zhang, Linfeng},
  journal       = {arXiv preprint arXiv:2506.06295},
  year          = {2025},
  eprint        = {2506.06295},
  archivePrefix = {arXiv}
}

@article{fu2025efficient,
  title         = {{Efficient-dLLM}: From Autoregressive to Diffusion Language Models, and Beyond in Speed},
  author        = {Fu, Yonggan and Whalen, Lexington and Ye, Zhifan and Dong, Xin and Diao, Shizhe and Liu, Jingyu and Wu, Chengyue and Zhang, Hao and Xie, Enze and Han, Song and others},
  journal       = {arXiv preprint arXiv:2512.14067},
  year          = {2025},
  eprint        = {2512.14067},
  archivePrefix = {arXiv}
}

@article{bie2025llada2-0,
  title         = {{LLaDA}2.0: Scaling Up Diffusion Language Models to 100B},
  author        = {Bie, Tiwei and Cao, Maosong and Chen, Kun and Du, Lun and Gong, Mingliang and Gong, Zhuochen and Gu, Yanmei and Hu, Jiaqi and Huang, Zenan and Lan, Zhenzhong and others},
  journal       = {arXiv preprint arXiv:2512.15745},
  year          = {2025},
  eprint        = {2512.15745},
  archivePrefix = {arXiv}
}

@article{bie2026llada2-1,
  title         = {{LLaDA}2.1: Speeding Up Text Diffusion via Token Editing},
  author        = {Bie, Tiwei and Cao, Maosong and Cao, Xiang and Chen, Bingsen and Chen, Fuyuan and Chen, Kun and Du, Lun and Feng, Daozhuo and Feng, Haibo and Gong, Mingliang and others},
  journal       = {arXiv preprint arXiv:2602.08676},
  year          = {2026},
  eprint        = {2602.08676},
  archivePrefix = {arXiv}
}

@inproceedings{song2026sparse,
  title     = {{Sparse-dLLM}: Accelerating Diffusion {LLMs} with Dynamic Cache Eviction},
  author    = {Song, Yuerong and Liu, Xiaoran and Li, Ruixiao and Liu, Zhigeng and Huang, Zengfeng and Guo, Qipeng and He, Ziwei and Qiu, Xipeng},
  booktitle = {Proceedings of the AAAI Conference on Artificial Intelligence},
  volume    = {40},
  pages     = {33038--33046},
  year      = {2026}
}

@inproceedings{miao2024specinfer,
  title     = {{SpecInfer}: Accelerating Large Language Model Serving with Tree-Based Speculative Inference and Verification},
  author    = {Miao, Xupeng and Oliaro, Gabriele and Zhang, Zhihao and Cheng, Xinhao and Wang, Zeyu and Zhang, Zhengxin and Wong, Rae Ying Yee and Zhu, Alan and Yang, Lijie and Shi, Xiaoxiang and others},
  booktitle = {Proceedings of the 29th ACM International Conference on Architectural Support for Programming Languages and Operating Systems, Volume 3},
  pages     = {932--949},
  year      = {2024}
}

@article{austin2021program,
  title         = {Program Synthesis with Large Language Models},
  author        = {Austin, Jacob and Odena, Augustus and Nye, Maxwell and Bosma, Maarten and Michalewski, Henryk and Dohan, David and Jiang, Ellen and Cai, Carrie and Terry, Michael and Le, Quoc and Sutton, Charles},
  journal       = {arXiv preprint arXiv:2108.07732},
  year          = {2021},
  eprint        = {2108.07732},
  archivePrefix = {arXiv}
}

@inproceedings{kwon2023pagedattention,
  title     = {Efficient Memory Management for Large Language Model Serving with {PagedAttention}},
  author    = {Kwon, Woosuk and Li, Zhuohan and Zhuang, Siyuan and Sheng, Ying and Zheng, Lianmin and Yu, Cody Hao and Gonzalez, Joseph and Zhang, Hao and Stoica, Ion},
  booktitle = {Proceedings of the ACM SIGOPS 29th Symposium on Operating Systems Principles},
  pages     = {611--626},
  year      = {2023}
}

@inproceedings{zheng2024sglang,
  title     = {{SGLang}: Efficient Execution of Structured Language Model Programs},
  author    = {Zheng, Lianmin and Yin, Liangsheng and Xie, Zhiqiang and Huang, Jeff and Sun, Chuyue and Yu, Cody Hao and Cao, Shiyi and Kozyrakis, Christos and Stoica, Ion and Gonzalez, Joseph E. and Barrett, Clark and Sheng, Ying},
  booktitle = {Advances in Neural Information Processing Systems},
  volume    = {37},
  year      = {2024}
}

@article{juravsky2024hydragen,
  title         = {{Hydragen}: High-Throughput {LLM} Inference with Shared Prefixes},
  author        = {Juravsky, Jordan and Brown, Bradley and Ehrlich, Ryan and Fu, Daniel Y. and R{\'e}, Christopher and Mirhoseini, Azalia},
  journal       = {arXiv preprint arXiv:2402.05099},
  year          = {2024},
  eprint        = {2402.05099},
  archivePrefix = {arXiv}
}

@inproceedings{bolya2023tome,
  title     = {Token Merging: Your {ViT} But Faster},
  author    = {Bolya, Daniel and Fu, Cheng-Yang and Dai, Xiaoliang and Zhang, Peizhao and Feichtenhofer, Christoph and Hoffman, Judy},
  booktitle = {International Conference on Learning Representations},
  year      = {2023}
}

@inproceedings{wang2021spatten,
  title     = {{SpAtten}: Efficient Sparse Attention Architecture with Cascade Token and Head Pruning},
  author    = {Wang, Hanrui and Zhang, Zhekai and Han, Song},
  booktitle = {2021 IEEE International Symposium on High-Performance Computer Architecture},
  pages     = {97--110},
  year      = {2021}
}

@inproceedings{rao2021dynamicvit,
  title     = {{DynamicViT}: Efficient Vision Transformers with Dynamic Token Sparsification},
  author    = {Rao, Yongming and Zhao, Wenliang and Liu, Benlin and Lu, Jiwen and Zhou, Jie and Hsieh, Cho-Jui},
  booktitle = {Advances in Neural Information Processing Systems},
  volume    = {34},
  pages     = {13937--13949},
  year      = {2021}
}

@inproceedings{xin2020deebert,
  title     = {{DeeBERT}: Dynamic Early Exiting for Accelerating {BERT} Inference},
  author    = {Xin, Ji and Tang, Raphael and Lee, Jaejun and Yu, Yaoliang and Lin, Jimmy},
  booktitle = {Proceedings of the 58th Annual Meeting of the Association for Computational Linguistics},
  pages     = {2246--2251},
  year      = {2020}
}

@inproceedings{schuster2022calm,
  title     = {Confident Adaptive Language Modeling},
  author    = {Schuster, Tal and Fisch, Adam and Gupta, Jai and Dehghani, Mostafa and Bahri, Dara and Tran, Vinh and Tay, Yi and Metzler, Donald},
  booktitle = {Advances in Neural Information Processing Systems},
  volume    = {35},
  pages     = {17456--17472},
  year      = {2022}
}

@techreport{fu2026nemotronlabsdiffusion,
  title       = {Nemotron-Labs-Diffusion: A Tri-Mode Language Model Unifying Autoregressive, Diffusion, and Self-Speculation Decoding},
  author      = {Yonggan Fu and Lexington Whalen and Abhinav Garg and Chengyue Wu and Maksim Khadkevich and Nicolai Oswald and Enze Xie and Daniel Egert and Sharath Turuvekere Sreenivas and Shizhe Diao and Chenhan Yu and Ye Yu and Weijia Chen and Sajad Norouzi and Jingyu Liu and Shiyi Lan and Ligeng Zhu and Jin Wang and Jindong Jiang and Morteza Mardani and Mehran Maghoumi and Song Han and Ante Jukic and Nima Tajbakhsh and Jan Kautz and Pavlo Molchanov},
  institution = {NVIDIA},
  year        = {2026},
  note        = {Technical report}
}

@article{wu2026blockbatch,
  title={BlockBatch: Multi-Scale Consensus Decoding for Efficient Diffusion Language Model Inference},
  author={Wu, Xiaoyou and Shih, Cheng-Jhih and Ji, Binfei and Liu, Yong and Lin, Yingyan Celine},
  journal={arXiv preprint arXiv:2605.29233},
  year={2026}
}
\bibliographystyle{iclr2027_conference}

% \newpage
% \clearpage
\appendix

\newpage

\section*{Appendix}
\begin{itemize}
  \item Sec.~\ref{appendix:foldedtransformer}: Folded Transformer Algorithm
    \begin{itemize}
      \item \hyperref[appendix:foldedattn]{Folded Self-attention}
      \item \hyperref[appendix:foldedffn]{Folded FFN}
      \item \hyperref[appendix:fused]{Putting it Together: One Folded Layer}
    \end{itemize}
  \item Sec.~\ref{appendix:triton}: Triton Implementation
  \item Sec.~\ref{appendix:numerical_discrepancy}: On the Numerical Discrepancy
        with Vanilla Decoding
  \item Sec.~\ref{app:additional_ablations}: Additional Ablations
    \begin{itemize}
      \item \hyperref[app:bscan_all]{Draft-budget Sweep across Model Scales and Families}
      \item \hyperref[app:depth_ablation]{Ablation on Drafting Strategy}
      \item \hyperref[app:gpu_ablation]{Ablation on Hardware}
    \end{itemize}

\end{itemize}

\section{Folded Transformer Algorithm}
\label{appendix:foldedtransformer}

This appendix develops the \textsc{SpecFoldLayer} in detail. Following the notations in the main text, given
the folded position set $\mathcal{R}^\ell_i$ and recomputed set
$\overline{\mathcal{R}}^\ell_i$ from Eq.~\ref{eq:specfold_sets}, we describe
how each Transformer sublayer is restructured to avoid redundant computation.
For brevity we drop the block index $k$ and layer index $\ell$, writing $\mathbf{H}_i$ for $\mathbf{H}^\ell_{i,k}$, $p$
for $\mathrm{par}(i)$, $\mathcal{R}_i$ for $\mathcal{R}^\ell_i$, and
$\overline{\mathcal{R}}_i$ for $\overline{\mathcal{R}}^\ell_i$ as we do a single layer's derivation.

\paragraph{Per-branch state.}
Within a layer, each branch $i$ owns three per-position tables over the $N$
block positions: (1) the RoPE-transformed projections $Q_i, K_i, V_i$, (2) the
streaming-softmax attention state $(m_i, Z_i, O_i)$ (per-row running maximum, denominator, and normalized output, as maintained by
FlashAttention), and (3) the FFN output table $F_i$. A branch initializes all
three tables by \emph{inheriting} its parent's and overwrites only the positions
in $\overline{\mathcal{R}}_i$ with recomputed values. The main branch $i{=}0$ has no parent and recomputes every position ($\overline{\mathcal{R}}_0 = [N]$). Crucially, the hidden states $\mathbf{H}_i$ themselves are never overwritten with parent copies, accumulating cross-branch
divergence.

The approximation enters in two places, both position-wise and gated by $\rho^\ell_{i,k,n} \le \delta$ (Eq.~\ref{eq:specfold_sets}). First, a folded position inherits
the parent's $(q, k, v)$ instead of projecting its own hidden state, approximating the self-attention while allowing computation reuse from parent (Sec.~\ref{appendix:foldedattn}). Second, a folded position reuses the parent's FFN output (Sec.~\ref{appendix:foldedffn}).

\subsection{Folded Self-attention}
\label{appendix:foldedattn}

The self-attention in DLLM is bidirectional and not position-wise. The output at position $n$ depends on every key and value in the sequence. Thus, a folded position has to correct its attention values regarding the recomputed positions. Therefore, \textsc{SpecFold} maintains, per head and per branch, the streaming-softmax state over the concatenated prefix and block KV caches, and updates the folded positions.

\paragraph{Setup.}
Let $(K_{\mathrm{P}}, V_{\mathrm{P}})$ be the prefix KV cache of length $P$,
shared by all branches, and let $q^i_n, k^i_n, v^i_n$ denote branch $i$'s
per-head table rows. Write
$s^i_{nj} = q^i_n \cdot k^i_j / \sqrt{d_h}$ for the score of query row $n$
against key $j$ of the concatenated table
$[K_{\mathrm{P}} \,\|\, K_i] \in \mathbb{R}^{(P+N) \times d_h}$. The
attention state of row $n$ is
\begin{equation}
    m_i[n] = \max_j s^i_{nj}, \qquad
    Z_i[n] = \sum_j e^{\, s^i_{nj} - m_i[n]}, \qquad
    O_i[n] = \frac{1}{Z_i[n]} \sum_j e^{\, s^i_{nj} - m_i[n]}\, v^i_j.
    \label{eq:flash_state}
\end{equation}

\paragraph{Recomputed query positions.}
For $n \in \overline{\mathcal{R}}_i$, the branch computes projections
from its own hidden state,
$(q^i_n, k^i_n, v^i_n) =
\mathrm{RoPE}\big(W_{QKV}\,\mathrm{RMSNorm}(\mathbf{H}_i[n])\big)$,
scatters them into its tables, and evaluates Eq.~\ref{eq:flash_state} with a
FlashAttention pass over $[K_{\mathrm{P}} \,\|\, K_i]$ and
$[V_{\mathrm{P}} \,\|\, V_i]$.

\paragraph{Folded query positions.}
For $n \in \mathcal{R}_i$, the position inherits the parent's projections,
\begin{equation}
    (q^i_n,\, k^i_n,\, v^i_n) \leftarrow (q^p_n,\, k^p_n,\, v^p_n),
    \label{eq:qkv_inherit}
\end{equation}
which is the sole attention-side approximation. The parent's state
$(m_p[n], Z_p[n], O_p[n])$ is the exact softmax attention of the (now
shared) query $q^p_n$ over the \emph{parent's} table. Since the child's
table differs from the parent's only on the positions in
$\overline{\mathcal{R}}_i$, the parent's state can be converted into the
child's exactly, by removing the parent's contributions at those positions and
adding the child's. With the rescaled maximum
$m' = \max\big(m_p[n],\,
\max_{e \in \overline{\mathcal{R}}_i} q^p_n \cdot k^i_e / \sqrt{d_h}\big)$
and $f = e^{\, m_p[n] - m'}$,
\begin{equation}
\begin{aligned}
    a_{\mathrm{old}}(e) &= e^{\, q^p_n \cdot k^p_e / \sqrt{d_h} \,-\, m'},
    \qquad
    a_{\mathrm{new}}(e) = e^{\, q^p_n \cdot k^i_e / \sqrt{d_h} \,-\, m'},
    \qquad e \in \overline{\mathcal{R}}_i, \\
    Z_i[n] &= Z_p[n]\, f
      \;-\; \textstyle\sum_{e}\, a_{\mathrm{old}}(e)
      \;+\; \textstyle\sum_{e}\, a_{\mathrm{new}}(e), \\
    O_i[n] &= \Big( O_p[n]\, Z_p[n]\, f
      \;-\; \textstyle\sum_{e}\, a_{\mathrm{old}}(e)\, v^p_e
      \;+\; \textstyle\sum_{e}\, a_{\mathrm{new}}(e)\, v^i_e \Big)
      \,\big/\, Z_i[n].
\end{aligned}
\label{eq:rank_correction}
\end{equation}

\subsection{Folded FFN}
\label{appendix:foldedffn}

After the attention state is assembled, the post-attention residual add is
formed for \emph{all} positions of every branch from the branch's own
hidden state:
\begin{equation}
    \mathbf{H}^+_i[n] = \mathbf{H}_i[n] + W_O\, O_i[n]
    \qquad \forall\, n \in \{0, \ldots, N{-}1\}.
    \label{eq:postattn_residual}
\end{equation}
As emphasized in Sec.~\ref{sec:sparspec:strategies}, the residual stream allows every position, whether recomputed or folded, to accumulate and preserve position-wise divergence.

The FFN sublayer is position-wise, so folded positions can reuse the
parent's output table directly:
\begin{equation}
    F_i[n] =
    \begin{cases}
        F_p[n], & n \in \mathcal{R}_i, \\[4pt]
        \mathrm{FFN}\big(\mathrm{RMSNorm}(\mathbf{H}^+_i[n])\big),
        & n \in \overline{\mathcal{R}}_i,
    \end{cases}
    \qquad\quad
    \mathbf{H}'_i[n] = \mathbf{H}^+_i[n] + F_i[n].
    \label{eq:ffn_reuse}
\end{equation}

Similar to the residual add at self-attention layer, $\mathbf{H}'_i[n]$ in Eq.~\ref{eq:ffn_reuse} retains the branch's
own hidden state, allowing the position to accumulate and preserve divergence.

\subsection{Putting it Together: One Folded Layer}
\label{appendix:fused}

Alg.~\ref{alg:specfoldlayer} composes the folded attention and folded FFN
into a single \textsc{SpecFoldLayer}. The branches are visited in topological order such that every branch inherits fully updated parent tables and applies the correction of Eq.~\ref{eq:rank_correction} against its immediate parent. The corrections propagate along fold chains, where the kernel implementation exploits this. It resolves
each folded row to its nearest recomputed ancestor and applies the
accumulated correction in one order-free launch over all branches (Appendix~\ref{appendix:triton}).

\begin{algorithm}[t]
\footnotesize
\caption{\textsc{SpecFoldLayer}: one transformer layer with inherited
QKV/FFN and exact incremental attention.}
\label{alg:specfoldlayer}
\begin{algorithmic}[1]
\Require Hidden states $\{\mathbf{H}_i\}_{i=0}^{B}$, parent map $\mathrm{par}(\cdot)$,
         recompute sets $\{\overline{\mathcal{R}}_i\}_{i=0}^{B}$,
         prefix KV cache $(K_{\mathrm{P}}, V_{\mathrm{P}})$ of length $P$,
         layer weights $(W_{QKV}, W_{O}, W_{\mathrm{FFN}})$, head dim $d_h$
\Ensure  Next-layer hidden states $\{\mathbf{H}'_i\}_{i=0}^{B}$
\For{each branch $i = 0, \ldots, B$ in topological order}
    \Comment{parents before children; kernel runs all branches in one order-free launch}
    \State $(Q_i, K_i, V_i) \leftarrow (Q, K, V)_{\mathrm{par}(i)}$;\quad
           $(m_i, Z_i, O_i) \leftarrow (m, Z, O)_{\mathrm{par}(i)}$;\quad
           $F_i \leftarrow F_{\mathrm{par}(i)}$
           \hfill\Comment{inherit parent tables}
    \For{$n \in \overline{\mathcal{R}}_i$}
        \Comment{recomputed positions: full QKV projection and attention}
        \State $(Q_i[n], K_i[n], V_i[n]) \leftarrow
               \mathrm{RoPE}\big(W_{QKV}\,\mathrm{RMSNorm}(\mathbf{H}_i[n])\big)$
        \State $(m_i[n], Z_i[n], O_i[n]) \leftarrow
               \mathrm{FlashAttn}\big(Q_i[n],\,
               [K_{\mathrm{P}} \,\|\, K_i],\, [V_{\mathrm{P}} \,\|\, V_i]\big)$
    \EndFor
    \For{$n \notin \overline{\mathcal{R}}_i$}
        \Comment{folded positions: correction of the inherited state}
        \State $m' \leftarrow \max\big(m_i[n],\,\max_{e\in\overline{\mathcal{R}}_i}Q_i[n]\!\cdot\!K_i[e]/\sqrt{d_h}\big)$;\quad
               $f \leftarrow e^{\,m_i[n]-m'}$
        \State $a_{\mathrm{old}}(e) \leftarrow
               e^{\,Q_i[n]\cdot K_{\mathrm{par}(i)}[e]/\sqrt{d_h} - m'}$,\;\;
               $a_{\mathrm{new}}(e) \leftarrow
               e^{\,Q_i[n]\cdot K_i[e]/\sqrt{d_h} - m'}$
               \textbf{ for } $e \in \overline{\mathcal{R}}_i$
        \State $Z' \leftarrow Z_i[n]\,f
               - \textstyle\sum_{e} a_{\mathrm{old}}(e)
               + \textstyle\sum_{e} a_{\mathrm{new}}(e)$
        \State $O' \leftarrow \big(O_i[n]\,Z_i[n]\,f
               - \textstyle\sum_{e} a_{\mathrm{old}}(e)\,V_{\mathrm{par}(i)}[e]
               + \textstyle\sum_{e} a_{\mathrm{new}}(e)\,V_i[e]\big)\,/\,Z'$
        \State $(m_i[n], Z_i[n], O_i[n]) \leftarrow (m', Z', O')$
    \EndFor
    \State $\mathbf{H}^{+}_i \leftarrow \mathbf{H}_i + O_i\,W_{O}$
           \hfill\Comment{residual never folded}
    \For{$n \in \overline{\mathcal{R}}_i$}
        \State $F_i[n] \leftarrow
               \mathrm{FFN}\big(\mathrm{RMSNorm}(\mathbf{H}^{+}_i[n])\big)$
               \hfill\Comment{folded positions reuse $F_{\mathrm{par}(i)}$}
    \EndFor
    \State $\mathbf{H}'_i \leftarrow \mathbf{H}^{+}_i + F_i$
\EndFor
\State \Return $\{\mathbf{H}'_i\}_{i=0}^{B}$
\end{algorithmic}
\end{algorithm}

\section{Triton Implementation}
\label{appendix:triton}

This section maps the \textsc{SpecFoldLayer} of Appendix~\ref{appendix:fused} onto the Triton implementation. The implementation runs six kernel families per layer, in order, over a flattened index space of $(B{+}1)N$
(branch, position) pairs to realize the \textsc{SpecFoldLayer} algorithm described in Appendix~\ref{appendix:foldedtransformer}.

\begin{table}[h]
\centering
\caption{Kernel pipeline for one \textsc{SpecFoldLayer}. $\overline{\mathcal{R}}$
denotes the union of the per-branch recompute sets
$\overline{\mathcal{R}}^\ell_i$. The grid sizes are per-layer launch dimensions.}
\label{tab:triton_kernels}
\small
\setlength{\tabcolsep}{4pt}
\begin{tabularx}{\linewidth}{@{}l l >{\raggedright\arraybackslash}X@{}}
\toprule
\textbf{Kernel} & \textbf{Grid} & \textbf{Role} \\
\midrule
\texttt{residual\_delta\_gate} & $(B{+}1)N$ & gate, source map, compaction, RMSNorm stats \\
\texttt{src\_resolve} & $(B{+}1)N$ & pointer-jump fold chains \\
\texttt{ln\_qkv\_rope\_scatter} & $|\overline{\mathcal{R}}|$ & norm + packed QKV + RoPE for recomputed positions \\
\texttt{flash\_attn} (+state) & $|\overline{\mathcal{R}}| \times H$ & recomputed-position attention; emits $(m, Z, O)$ \\
\texttt{chain\_correction} & $(B{+}1)N \times H$ & correcting attention for folded positions \\
\texttt{ln\_gateup\_silu} / & $|\overline{\mathcal{R}}|$ & fused FFN + residual add for recomputed positions \\
\quad\texttt{down\_residual\_scatter} & & \\
\bottomrule
\end{tabularx}
\end{table}

\paragraph{Gate and compaction.}
One program per (branch, position) pair evaluates the relative residual of
Eq.~\ref{eq:s2_sim} against the parent's corresponding position. Each program writes three outputs: a recompute flag, a source index (self if recomputed, the parent otherwise), and, via atomic counters, its flat index into two compacted lists: a global recomputed position list that the GEMM and attention kernels iterate over, and a per-branch list that the correction kernel uses to visit one branch's recomputed entries. The main
branch bypasses the gate and is marked recompute at every position.

\paragraph{Source resolution.}
The sources form chains through the draft graph as a folded position can point at a position that the corresponding position at parent is also folded. A source-resolution kernel follows each chain hop by hop, after which every folded position addresses the nearest ancestor in which the position is recomputed.

\paragraph{Projections.}
Recomputed positions are processed as one contiguous batch through a kernel
that applies RMSNorm, multiplies by the
QKV projection weights packed into a single matrix, applies rotary embeddings, and scatters the results into the full $Q$, $K$, $V$ tables at their (branch, position) slots. Folded slots stay untouched.

\paragraph{Attention with state reuse.}
A FlashAttention-style kernel processes only recomputed query rows, streaming over the shared prefix cache and over the in-block KV tables. Besides the attention output, it retains each row's running softmax maximum $m$ and denominator $Z$ in per-head state tables. A second kernel serves folded positions: starting from the source row's
$(m, Z, O)$, it walks the fold chain hop by hop and applies the exact
correction of Alg.~\ref{alg:specfoldlayer}, swapping the ancestor's key and
value contributions for the current branch's recomputed ones at that branch's
segmented entries only.

\paragraph{FFN and residual.}
A kernel applies RMSNorm, the gate and up projections packed into one GEMM, the SiLU product, and the down projection for recomputed rows, scattering the FFN outputs into a table. Every position then adds its own residual stream, which is never folded, to its FFN output.

The \textsc{SpecFoldLayer} implementation is in Triton over HuggingFace checkpoints for both evaluated model families. Embeddings and the language-model head remain in PyTorch or cuBLAS libraries.

\section{On the Numerical Discrepancy with Vanilla Decoding}
\label{appendix:numerical_discrepancy}
In principle, the decode outcome of multi-branch speculative decoding in DLLM is equivalent to vanilla DLLM decoding. Every accepted draft token is, by construction, the same token vanilla decoding would have produced from the same context (Sec.~\ref{sec:background:speculative}). With this theory, the Spiffy~\cite{agrawal2025spiffy}
row in our metric tables should  match the Vanilla baseline.
In practice, we observe small discrepancies across
benchmarks. This appendix explains the source of this discrepancy and describes
the sanity check we ran to confirm that our DLLM speculative decoding
implementation is correct.

The source of this discrepancy is introduced by numerical drifts in tensor operator libraries. As also documented by Spiffy~\cite{agrawal2025spiffy}, \texttt{torch.bfloat16} matrix multiplication (\texttt{matmul})
on CUDA GPUs can use an accumulation order that depends on the shape of the
input tensor. Once speculative decoding batches the main branch together with $B$ draft branches, the \texttt{matmul} reduces over a tensor whose
leading dimension differs from the Vanilla case, where no speculative draft is
present ($B{=}0$). Even though the draft branches are independent and
contribute no information to the main branch's logits in exact arithmetic, the
different accumulation order produces rounding differences in the last few
bits of each \texttt{matmul} output. These differences are amplified by the short
mantissa field of \texttt{bfloat16}. Over many denoising steps, they can
occasionally flip a token at a position where two candidates have nearly equal
logits. Since exact match and pass@1 are non-smooth metrics, a single flipped
token can change a sample-level score, yielding the inevitable small accuracy drift we
observe.

To justify that the discrepancy is caused by batched \texttt{matmul} numerics rather than an algorithmic implementation error, we ran a sanity check in which the $B{+}1$ branches are forwarded through the DLLM one at a time rather than as a single batched forward pass. With $B{=}0$ for every forward pass, the \texttt{matmul} shapes match the Vanilla DLLM forward pass, so the accumulation order is preserved. In this serialized run, our implementation reproduces the Vanilla baseline metrics on every benchmark we tested. This rules out an algorithmic implementation error in the speculative verification logic.

\section{Additional Ablations}
\label{app:additional_ablations}

This section presents the additional ablations conducted for the \textsc{SpecFold} algorithm and implementation.

\subsection{Draft-budget sweep across model scales and families}
\label{app:bscan_all}

\begin{figure}[h]
    \centering
    \includegraphics[width=\linewidth]{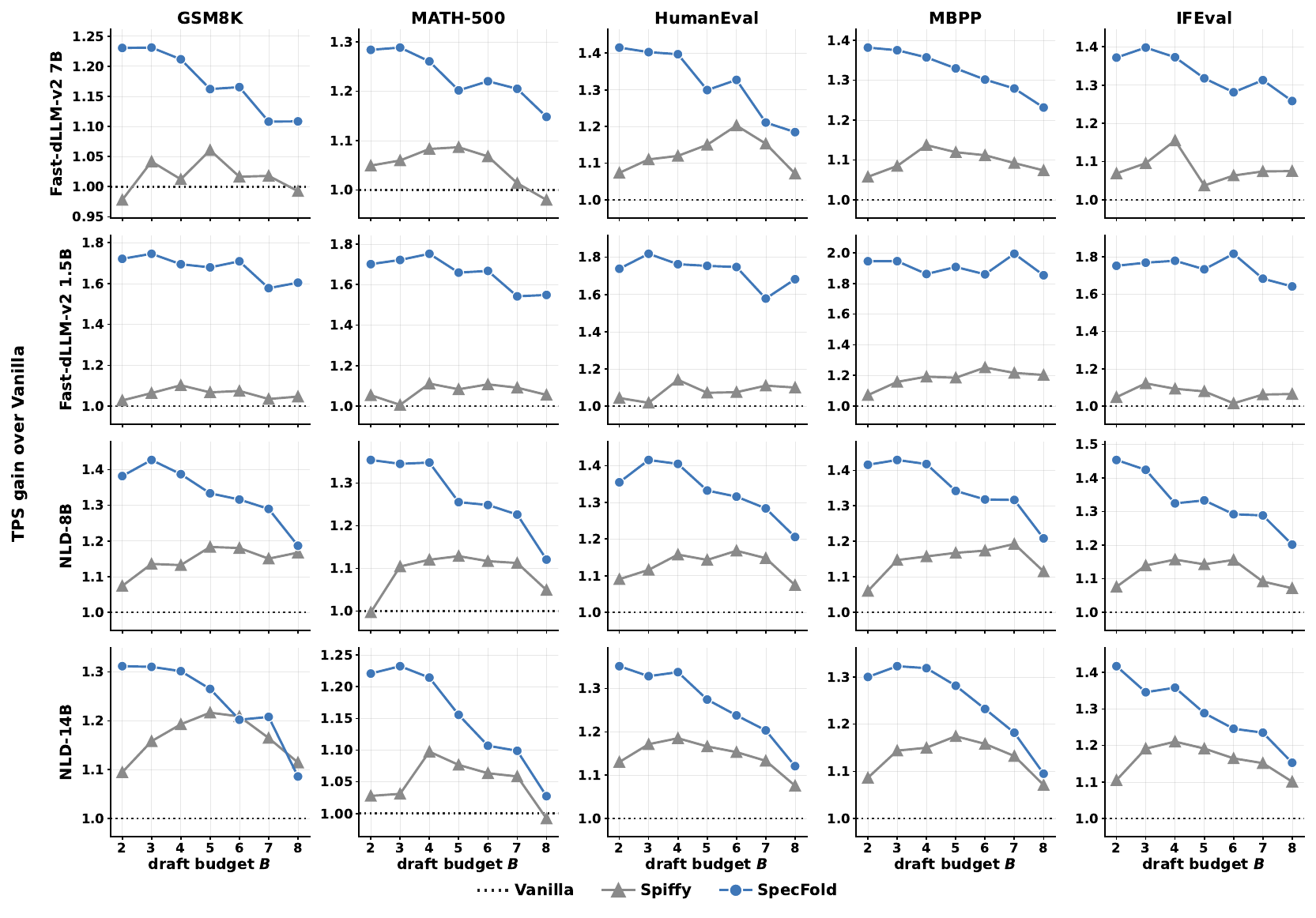}
    \vspace{-1.8em}
    \caption{
    Draft-budget sweep on the four remaining models (rows) across the five
    benchmarks (columns), presented as in Fig.~\ref{fig:bscan_nld3b}: each
    panel plots the throughput gain (TPS) of Spiffy and \textsc{SpecFold}
    over Vanilla against the draft budget $B$, with the dotted line marking
    Vanilla. At every $B$ the two methods share identical drafts and NFE,
    so the gap between the curves is the effect of folding.
    }
    \vspace{-0.5em}
    \label{fig:bscan_all_models}
\end{figure}

We extend the draft-budget sweep of Sec.~\ref{sec:ablation} from
Nemotron-Labs-Diffusion-3B to the four remaining models of
Table~\ref{tab:h100_all} under the identical setup.
Fig.~\ref{fig:bscan_all_models} collects all four models in one grid. 

Across models and benchmarks, \textsc{SpecFold} sustains a consistent throughput advantage over Spiffy. As the draft budget $B$ grows, however, draft acceptance saturates while every added branch still adds verification cost. Therefore, the end-to-end throughput gain diminishes toward the dense baseline.

\subsection{Ablation on Drafting Strategy}
\label{app:depth_ablation}

\begin{figure}[h]
    \centering
    \includegraphics[width=\linewidth]{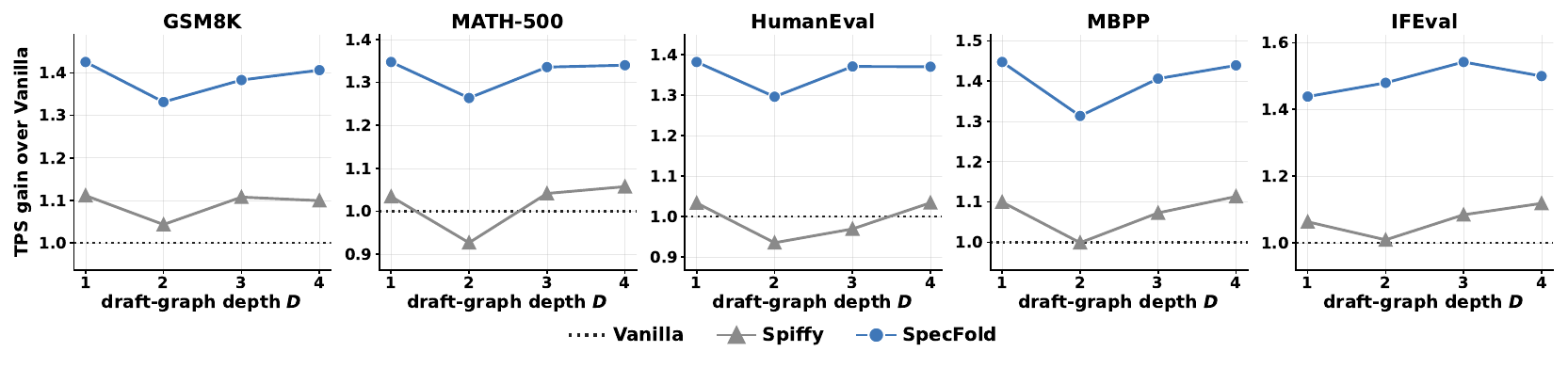}
    \vspace{-1.8em}
    \caption{
    Drafting-strategy ablation on Nemotron-Labs-Diffusion-3B: throughput
    gain (TPS) over Vanilla versus the draft-graph depth cap $D$ at fixed
    budget $B{=}7$, presented as in Fig.~\ref{fig:bscan_nld3b}. Spiffy and
    \textsc{SpecFold} share identical drafts and NFE at every depth.
    }
    \vspace{-0.5em}
    \label{fig:depth_ablation}
\end{figure}

We extend our investigation in 
\textsc{SpecFold}'s effectiveness on the choice of shape of draft graph and drafting strategy. We constrain the drafting
strategy by capping the draft-graph depth at $D \in \{1,2,3,4\}$ on
Nemotron-Labs-Diffusion-3B, following settings in Table~\ref{tab:h100_all}. This setup spans the two extremes of drafting strategy: $D{=}1$ yields seven sibling drafts that each differ from the main branch by a single token, and $D{=}4$ yields deep chains in which each draft extends its parent. Fig.~\ref{fig:depth_ablation} plots the throughput gain (TPS) over Vanilla for both methods at every depth.

We demonstrate that \textsc{SpecFold}'s advantage is insensitive to the drafting strategy. It
delivers a $1.26$--$1.54\times$ throughput gain over Vanilla at every depth on every
benchmark, with fold ratios of $0.52$--$0.67$ across all shapes. This demonstrates that, while drafting impacts the acceptance of speculative drafts, \textsc{SpecFold} provides consistent throughput gains.

\subsection{Ablation on Hardware}
\label{app:gpu_ablation}

To test whether \textsc{SpecFold}'s gains depend on a particular accelerator,
we repeat the Nemotron-Labs-Diffusion-3B configuration of the sweeps above
(recompute threshold $\delta=0.1$, $B=8$ draft branches, identical
calibration graphs) on NVIDIA A100 40GB PCIe GPUs and compare against H100 runs of
the same configuration.

\begin{table}[h]
\centering
\caption{GPU-type ablation on Nemotron-Labs-Diffusion-3B: decoding throughput (tokens/s), TPS gain over the same GPU's vanilla decoding, and task accuracy.}
\label{tab:gpu_ablation}
\small
\setlength{\tabcolsep}{4pt}
\begin{tabular}{llccc|ccc}
\toprule
\multirow{2}{*}{\textbf{Benchmark}} & \multirow{2}{*}{\textbf{Method}}
& \multicolumn{3}{c|}{\textbf{A100}} & \multicolumn{3}{c}{\textbf{H100}} \\
\cmidrule(lr){3-5}
\cmidrule(lr){6-8}
& & \textbf{TPS$\uparrow$} & \textbf{TPS gain} & \textbf{Acc.$\uparrow$}
& \textbf{TPS$\uparrow$} & \textbf{TPS gain} & \textbf{Acc.$\uparrow$} \\
\midrule
\multirow{3}{*}{GSM8K}
& Vanilla & 81 & 1.00$\times$ & 88.32\% & 144 & 1.00$\times$ & 88.02\% \\
& Spiffy & 92 & 1.14$\times$ & 88.02\% & 162 & 1.12$\times$ & 87.95\% \\
\rowcolor{colOurs}
& \textbf{\textsc{SpecFold}} & 121 & \textbf{1.49$\times$} & 88.78\% & 197 & 1.36$\times$ & 88.63\% \\
\midrule
\multirow{3}{*}{MATH}
& Vanilla & 92 & 1.00$\times$ & 72.80\% & 156 & 1.00$\times$ & 72.60\% \\
& Spiffy & 103 & 1.12$\times$ & 74.20\% & 167 & 1.07$\times$ & 72.60\% \\
\rowcolor{colOurs}
& \textbf{\textsc{SpecFold}} & 131 & \textbf{1.43$\times$} & 75.00\% & 197 & 1.26$\times$ & 73.20\% \\
\midrule
\multirow{3}{*}{HumanEval}
& Vanilla & 62 & 1.00$\times$ & 73.78\% & 110 & 1.00$\times$ & 73.78\% \\
& Spiffy & 75 & 1.20$\times$ & 74.39\% & 114 & 1.04$\times$ & 75.61\% \\
\rowcolor{colOurs}
& \textbf{\textsc{SpecFold}} & 100 & \textbf{1.60$\times$} & 72.56\% & 145 & 1.32$\times$ & 71.34\% \\
\midrule
\multirow{3}{*}{MBPP}
& Vanilla & 65 & 1.00$\times$ & 70.37\% & 113 & 1.00$\times$ & 73.02\% \\
& Spiffy & 75 & 1.15$\times$ & 69.84\% & 125 & 1.11$\times$ & 70.37\% \\
\rowcolor{colOurs}
& \textbf{\textsc{SpecFold}} & 100 & \textbf{1.55$\times$} & 70.63\% & 153 & 1.35$\times$ & 73.54\% \\
\midrule
\multirow{3}{*}{IFEval}
& Vanilla & 54 & 1.00$\times$ & 59.52\% & 94 & 1.00$\times$ & 60.26\% \\
& Spiffy & 66 & 1.23$\times$ & 60.07\% & 98 & 1.05$\times$ & 60.63\% \\
\rowcolor{colOurs}
& \textbf{\textsc{SpecFold}} & 90 & \textbf{1.67$\times$} & 59.33\% & 139 & 1.48$\times$ & 59.15\% \\
\bottomrule
\end{tabular}
\end{table}

As shown in Table~\ref{tab:gpu_ablation}, \textsc{SpecFold}'s relative gain is
larger on A100 ($1.43$--$1.67\times$) than on H100
($1.26$--$1.48\times$). Task accuracy is preserved. On the A100, which has lower compute capacity, the dense draft batch accounts for a larger share of each forward pass, so \textsc{SpecFold} recovers more throughput. In contrast, the Spiffy baseline, which pays the dense forward pass cost for every draft, stays within $1.04$--$1.23\times$ TPS gain on both GPUs. 

\end{document}